\documentclass[journal]{IEEEtran}

\IEEEoverridecommandlockouts

\usepackage{amsmath,amssymb,amsfonts,amsthm,dsfont} 
\usepackage{algorithm,algorithmicx,listings}  
\usepackage[noend]{algpseudocode}			  
\usepackage{hyperref}
\usepackage{graphics,graphicx,subfigure,color}	
\usepackage{enumitem, ulem}	 
\usepackage{array,multirow,multicol,rotating,slashbox}			
\usepackage{gensymb}		
\usepackage[table, usenames, dvipsnames]{xcolor}

\usepackage{extarrows}

\def\argmin{\mathop{\arg\,\min}\limits}	%
\def\argmax{\mathop{\arg\,\max}\limits}	%

\newcommand{\tcn}[1]{\cellcolor{Magenta!#1!TealBlue}#1}

\newtheorem*{example*}{Example}
\newtheorem*{problem*}{Problem}

\begin{document}

\title{Nonmyopic View Planning for Active Object Detection}

\author{Nikolay~Atanasov,~\IEEEmembership{Student Member,~IEEE,}
        Bharath~Sankaran,~\IEEEmembership{Student Member,~IEEE,} 
        Jerome~Le~Ny,~\IEEEmembership{Member,~IEEE,}
        George~J.~Pappas,~\IEEEmembership{Fellow,~IEEE,}
        and~Kostas~Daniilidis,~\IEEEmembership{Fellow,~IEEE}
\thanks{This work was supported in part by TerraSwarm, one of six centers of STARnet, a Semiconductor Research Corporation program sponsored by MARCO and DARPA. It was supported also by the NSF-IIP-0742304, NSF-OIA-1028009, ARL MAST-CTA W911NF-08-2-0004, ARL RCTA W911NF-10-2-0016, and NSF-DGE-0966142 grants.}
\thanks{N. Atanasov and G. J. Pappas are with the department of Electrical and Systems Engineering, University of Pennsylvania, Philadelphia, PA 19104, {\tt\small\{atanasov, pappasg\}@seas.upenn.edu}.}%
\thanks{B. Sankaran is with the Computational Learning and Motor Control Lab, University of Southern California, Los Angeles, CA 90089, {\tt\small bsankara@usc.edu}}%
\thanks{J. Le Ny is with the department of Electrical Engineering, Ecole Polytechnique de Montreal, QC, Canada, {\tt\small jerome.le-ny@polymtl.ca}}%
\thanks{K. Daniilidis is with the department of Computer and Information Science, University of Pennsylvania, Philadelphia, PA 19104, {\tt\small kostas@cis.upenn.edu}}%
\thanks{This work has been submitted to the IEEE for possible publication. Copyright may be transferred without notice, after which this version may no longer be accessible.}%
}


\maketitle

\begin{abstract}
One of the central problems in computer vision is the detection of semantically important objects and the estimation of their pose. Most of the work in object detection has been based on single image processing and its performance is limited by occlusions and ambiguity in  appearance and geometry. This paper proposes an active approach to object detection by controlling the point of view of a mobile depth camera. When an initial static detection phase identifies an object of interest, several hypotheses are made about its class and orientation. The sensor then plans a sequence of views, which balances the amount of energy used to move with the chance of identifying the correct hypothesis. We formulate an active hypothesis testing problem, which includes sensor mobility, and solve it using a point-based approximate POMDP algorithm. The validity of our approach is verified through simulation and real-world experiments with the PR2 robot. The results suggest that our approach outperforms the widely-used greedy view point selection and provides a significant improvement over static object detection.
\end{abstract}




\IEEEpeerreviewmaketitle
\section{Introduction}
\label{sec:intro}
\IEEEPARstart{W}{ith} the rapid progress of robotics research, the utility of autonomous robots is no longer restricted to controlled industrial environments. The focus has shifted to high-level interactive tasks in complex environments. The effective execution of such tasks requires the addition of semantic information to the traditional traversability representation of the environment. For example, household robots need to identify objects of interest and estimate their pose accurately in order to perform manipulation tasks.

One of the central problems in computer vision, object detection and pose estimation, historically has been addressed with the assumption that the pose of the sensing device is fixed \cite{Lowe04}, \cite{BelongieMP02}, \cite{FanMN89}. However, occlusions, variations in lighting, and imperfect object models in realistic environments decrease the accuracy of single-view object detectors. Active perception approaches circumvent these issues by utilizing appropriate sensing settings to gain more information about the scene. A large body of research in \textit{sensor management} \cite{Hero11_sensor_management} presents a structured approach to controlling the degrees of freedom in sensor systems in order to improve the results of the information acquisition process. However, most of the work either assumes a simplified model for the detection process \cite{spaan08_POMDP, Jenkins10_PhD} or avoids the problem altogether and concentrates on estimating a target's state after its detection \cite{Hero03_sensor_management, sommerlade08_information, Mihaylova03_active_sensing}.

This paper is a step towards bridging the gap between the research in sensor management and the recent advances in 3D object detection enabled by the advent of low-cost RGB-D cameras and open-source point cloud libraries \cite{Rusu_ICRA2011_PCL}. Rather than placing the burden of providing perfect results on a static detector, the sensor can move to increase the confidence in its detection. We consider the following problem. A mobile sensor has access to the models of several objects of interest. Its task is to determine which, if any, of the objects of interest are present in a cluttered scene and to estimate their poses. The sensor has to balance the detection accuracy with the time spent observing the objects. The problem can be split into a static detection stage followed by a planning stage to determine a sequence of points of view, which minimize the mistakes made by the observer.

A preliminary version of this paper appeared at the 2013 IEEE International Conference of Robotics and Automation \cite{atanasov13_active_object_detection}. It included the problem formulation, theoretical development, and an initial evaluation of our approach in simulation. This version clarifies the theoretical development, provides extensive simulation results, and demonstrates the performance in real-world experiments using an Asus Xtion RGB-D sensor attached to the wrist of a PR2 robot.   

The rest of the paper is organized as follows. The next section provides an overview of the related approaches to active perception and summarizes our contribution. In Section \ref{sec:problem} we draw up hypotheses about the class and orientation of an unknown object and formulate the active detection problem precisely. Section \ref{sec:obj_detect} presents our approach to static detection using a depth camera. In Section \ref{sec:obs_model} we discuss the generation of an observation model, which is used to assign a confidence measure to the hypotheses in a Bayesian framework. Section \ref{sec:hyp_test} describes how to choose a sequence of views for the sensor, which is used to test the hypotheses and balances sensing time with decision accuracy. Implementation details are discussed in Section \ref{sec:impl_det}. Finally, in Section \ref{sec:experiments} we present results from simulation and real-world experiments and discuss the validity of our approach.

\section{Related Work}
\label{sec:rel_work}
The approaches to active detection and estimation in sensor management \cite{Hero11_sensor_management, Huber09_PhD} can be classified according to sensor type into \textit{mobile} (sensors have dynamic states) and \textit{stationary} (sensors have fixed states). The objective may be to identify a target and estimate its state or simply to improve the state estimate of a detected target. The targets might be mobile or stationary as well. The process of choosing sensor configurations may be \textit{myopic} (quantifies the utility of the next configuration only) or \textit{non-myopic} (optimizes over a sequence of future configurations).

Some of the earliest work in active perception is done by Bajscy \cite{bajscy_88, Krotkov_93}. It is focused on 3D position estimation through control of the sensor's intrinsic parameters. Pito's paper \cite{pito99_nbv} addresses the next best view problem for surface reconstruction as one that maximizes information gain by increasing spatial resolution. The movement of the sensor is constrained to a circle centered around the object of interest.

The work that is closest to ours \cite{Scharinger10_SceneModeling6D} uses a mobile sensor to classify stationary objects and estimate their poses. Static detection is performed using SIFT matching. An object's pose distribution is represented with a Gaussian mixture. A myopic strategy is used to reduce the differential entropy in the pose and class distributions. This work differs from ours in that the sensor has models of all the objects so the detection is never against background. Moreover, our approach is non-myopic.

Velez and coworkers \cite{velez11_ijcai, velez11_icaps} consider detecting doorways along the path of a mobile sensor traveling towards a fixed goal. The unknown state of a candidate detection is binary: ``present'' or ``not present''. A method due to \cite{Felzenszwalb08_detector} is used for static detection. Stereo disparity and plane fitting are used for pose estimation. An entropy field is computed empirically for all viewpoints in the workspace and is used to myopically select locations with high expected information gain. The authors assume that the static detector provides accurate pose estimates and do not optimize them during the planning. In our work we use a depth sensor, which validates the assumption that position estimates need not be optimized. However, the orientation estimates can be improved through active planning. Inspired by the work on hypothesis testing \cite{Javidi12_arxiv}, we introduce a rough discretization of the space of orientations so that the hidden object state takes on several values, one for ``object not present'' and the rest for ``object present'' with a specific orientation. In our previous work we considered a dual hypothesis problem aimed at model completion \cite{BharathThesis}.

Karasev et al. \cite{karasev12_visual_learning} plan the path of a mobile sensor for visual search of an object in an otherwise known and static scene. The problem statement is different from ours but the optimization is surprisingly similar. The authors hypothesize about the pose of the object and use an analytical model of the sensing process to myopically maximize the conditional entropy of the next measurement.

A lot of the work in sensor management assumes a fixed sensor position, which simplifies the problem considerably because the trade-off between minimizing movement energy and maximizing viewpoint informativeness is avoided \cite{sommerlade08_information, Kragic06_ActiveObjRecognition}. Frequently, the action selection process is myopic. For example, the authors of \cite{sommerlade08_information, Kragic06_ActiveObjRecognition} control a pan-zoom-tilt camera with a fixed position in order to track mobile targets based on greedy minimization of conditional entropy. In contrast, we consider a mobile sensor, include the detection process in the optimization, and use non-myopic planning. Golovin and Krause \cite{golovin11_adaptive} showed that myopic planning for an adaptively submodular objective function is merely by a constant factor worse than the optimal strategy. Unfortunately, the objective in our formulation is not adaptively submodular and even with a fixed sensor state, a myopic strategy can perform arbitrarily worse than the optimal policy \cite{Javidi12_arxiv}.

The contributions of the paper are as follows. First, we introduce implicit pose estimation in 3D object detection via the viewpoint-pose tree (VP-Tree). The VP-Tree is a static detector based on partial view matching, which provides a coarse pose estimate in addition to the detection of an object's class. Relying on partial views also helps in cases when the object of interest is partially occluded or in contact with another object. Second, we introduce a hypothesis testing approach to improve upon the static detection results by moving the sensor to more informative viewpoints. Our non-myopic plan weights the benefit of gaining more certainty about the correct hypothesis against the physical cost of moving the sensor. 


\section{Problem Formulation}
\label{sec:problem}
\subsection{Sensing}
\begin{figure*}[thb!]
	\includegraphics[width=0.7\linewidth, trim=55mm 0mm 43mm 12mm,clip]{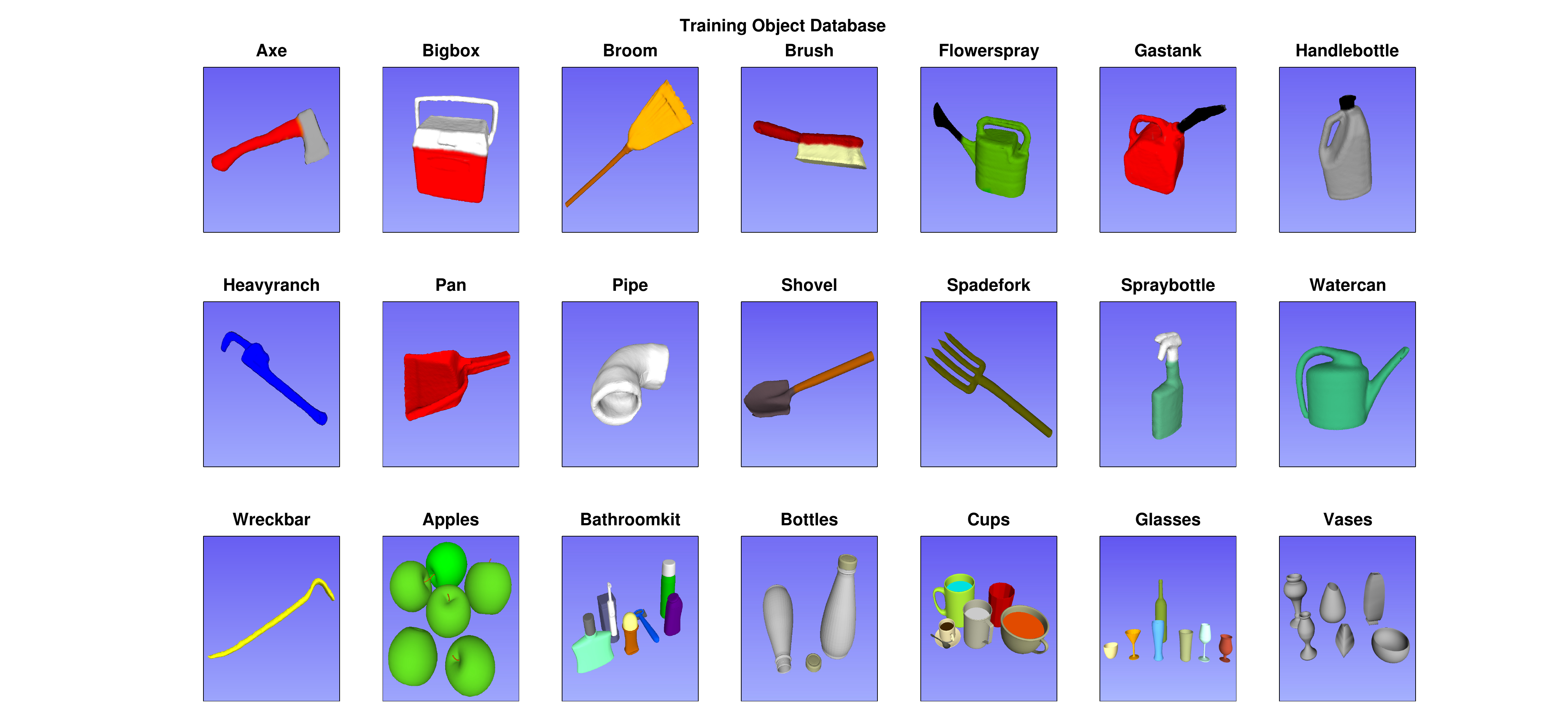}
	\includegraphics[width=0.29\linewidth]{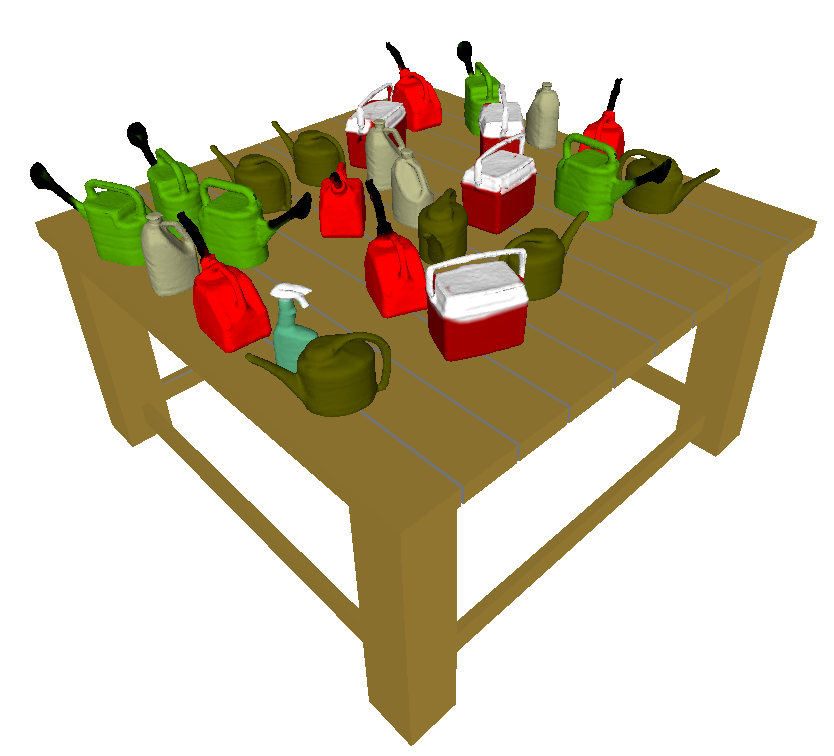}
	\caption{Database of object models constructed using kinect fusion \cite{Rusu_ICRA2011_PCL} (left) and an example of a scene used to evaluate our framework in simulation (right).}
	\label{fig:Database}
\end{figure*}

Consider a mobile depth sensor, which observes a \textit{static} scene, containing unknown objects. The sensor has access to a finite database $\mathcal{D}$ of object models (Fig. \ref{fig:Database}) and a subset $\mathcal{I}$ of them are designated as objects of interest. We assume that an object class has a single model associated with it and use the words model and class interchangeably. This is necessary because our static detector, the VP-Tree, works with instances. However, our approach for choosing an informative sensing path is independent of the static detector and can be used with class-based detectors.

The task of the sensor is to detect all objects from $\mathcal{I}$, which are present in the scene and to estimate their pose as \textit{quickly} as possible. Note that the detection is against not only known objects from the database but also clutter and background. At each time step the sensor obtains a point cloud from the scene, splits it into separate surfaces (\textit{segmentation}) and associates them with either new or previously seen objects (\textit{data association}). These procedures are not the focus of the paper but we mention how we perform them in Subsection \ref{subsec:seg_data_ass}. We assume that they estimate the object positions accurately.

We formulate hypotheses about the class and orientation of an unknown object by choosing a \textit{small} finite set of discrete orientations $\mathcal{R}(c) \subset SO(3)$ for each object class $c \in \mathcal{I}$. To denote the possibility that an object is not of interest we introduce a dummy class $c_\emptyset$ and a dummy orientation $\mathcal{R}(c_\emptyset) = \{r_\emptyset\}$. The sensor needs to decide among the following hypotheses:
\begin{align*}
H(c_\emptyset,r_\emptyset):&\text{ the object does not belong to $\mathcal{I}$,}\\
H(c,r):&\text{ the object class is $c \in \mathcal{I}$ with orientation $r \in \mathcal{R}(c)$}
\end{align*}
In order to measure the correctness of the sensor's decisions we introduce a cost for choosing $H(\hat{c},\hat{r})$, when $H(c,r)$ is correct:
\begin{align*}
J_D(\hat{c},\hat{r},c,r) := \begin{cases}
  		K(\hat{r}, r), & \hat{c} = c\\
  		K_+, & \hat{c} \in \mathcal{I}, c \notin \mathcal{I}\\
  		K_-, & \hat{c} \neq c, c \in \mathcal{I},
   \end{cases}
\end{align*}
where $K_+$ and $K_-$ are costs for making false positive and false negative mistakes respectively, and $K(\cdot,\cdot)$ is a cost for an incorrect orientation estimate, when the class is correct.

\begin{example*}
Suppose that the task is to look for chairs ($c_1$) and tables ($c_2$) regardless of orientation ($K(\hat{r}, r) := 0$). The decision cost can be represented with the matrix:
\[
J_D(\hat{c},\hat{r},c,r): \quad \begin{tabular}{l|c|c|c|}
\backslashbox{$\hat{c}$}{$c$} & $c_\emptyset$ & $c_1$ & $c_2$\\\hline
$c_\emptyset$ & $0$ & $K_-$ & $K_-$\\\hline
$c_1$ & $K_+$ & $0$ & $K_-$\\\hline
$c_2$ & $K_+$ & $K_-$ & $0$\\\hline
\end{tabular}
\]
In static detection, it is customary to run a chair classifier first to distinguish between $c_\emptyset$ and $c_1$ and then a table classifier to distinguish between $c_\emptyset$ and $c_2$. Our framework requires moving the sensor around the object to distinguish among the hypotheses and it is necessary to process them concurrently. 
\end{example*}

\subsection{Mobility}
We are interested in choosing a sequence of views for the mobile sensor, which has an optimal trade-off between the energy used to move and the expected cost of incorrect decisions. Doing this with respect to all objects in the scene simultaneously results in a complex joint optimization problem. Instead, we treat the objects independently and process them sequentially, which simplifies the task to choosing a sequence of sensor poses to observe a single object. Further, we restrict the motion of the sensor to a sphere of radius $\rho$, centered at the location of the object. The sensor's orientation is fixed so that it points at the centroid of the object. We denote this space of sensor poses by $V(\rho)$ and refer to it as a \textit{viewsphere}. A sensor pose $x \in V(\rho)$ is called a \textit{viewpoint}. See Fig. \ref{fig:GastankViewsphere} for an example. At a high-level planning stage we assume that we can work with a fully actuated model of the sensor dynamics. The viewsphere is discretized into a set of viewpoints $\mathcal{X}(\rho)$, described by the nodes of a graph. The edges connect nodes which are reachable within a single time step from the current location based on the kinematic restrictions of the sensor. Since the motion graph is known a priori the Floyd-Warshall algorithm can be used to precompute the all-pairs movement cost between viewpoints:
\begin{align*}
g(x,x') &= g_M(x,x') + g_0 = \text{ cost of moving from $x$ to $x'$ on}\\
&\text{ the viewsphere $\mathcal{X}(\rho)$ and taking another observation},
\end{align*}
where $g_0 > 0$ is a fixed measurement cost, which prevents the sensor from obtaining an infinite number of measurements without moving. As a result, a motion plan of length $T$ for the sensor consists of a sequence of viewpoints $x_1,\ldots,x_T \in \mathcal{X}(\rho)$ on the graph and its cost is:
\[
J_M(T) := \sum_{t=2}^T g(x_{t-1}, x_t) 
\]

\subsection{Active Object Detection}

\begin{problem*}
\label{prob:active_obj_detect}
Let the initial pose of the mobile sensor be $x_1 \in \mathcal{X}(\rho)$. Given an object with unknown class $c$ and orientation $r$, choose a stopping time $\tau$, a sequence of viewpoints $x_2, \ldots, x_\tau \in \mathcal{X}(\rho)$, and a hypothesis $H(\hat{c},\hat{r})$, which minimize the total cost:
\begin{align}
\label{eq:sequential_optimization}
\mathbb{E} \biggl\{ J_M(\tau) + \lambda J_D(\hat{c},\hat{r},c,r)\biggr\},
\end{align}
where $\lambda \geq 0$ determines the relative importance of a correct decision versus cost of movement. The expectation is over the the correct hypothesis and the observations collected by the sensor.
\end{problem*}

Our approach to solving the active object detection problem consists of two stages. First, we use a VP-Tree to perform static detection in 3D as described in the next section. Since the detection scores are affected by noise and occlusions, they are not used directly. Instead, the hypotheses about the detection outcome are maintained in a probabilistic framework. In the second stage, we use non-myopic planing to select better viewpoints for the static detector and update the probabilities of the hypotheses.

\section{Static Object Detection}
\label{sec:obj_detect}
In this section we introduce the VP-Tree, a static object detector, which is built on the principles of the vocabulary tree, introduced by Nister and Stewenius \cite{NisterS06_vocabtree}. A vocabulary tree is primarily used for large scale image retrieval where the number of semantic classes is in the order of a few thousand. The VP-Tree extends the utility of the vocabulary tree to joint recognition and pose estimation in 3D by using point cloud templates extracted from various viewpoints around the object models in the database $\mathcal{D}$. The templates serve to discretize the orientation of an object and make it implicit in the detection. Given a query point cloud, the best matching template carries information about both the class and the pose of the object relative to the sensor.

\begin{figure}[t!]
	\begin{center}
		\includegraphics[width=\linewidth]{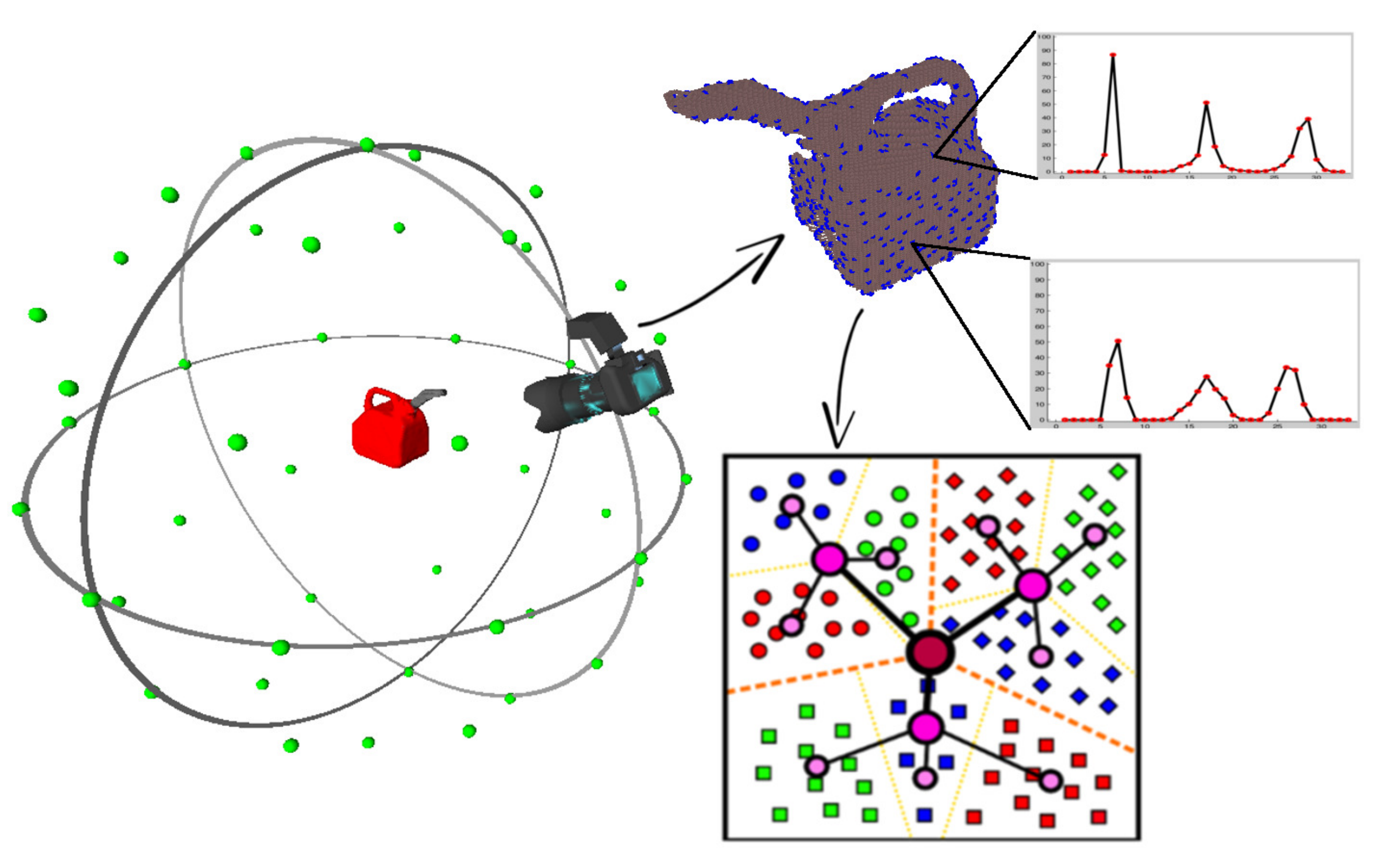}
	\end{center}
	\caption{The sensor position is restricted to a set of points on a sphere centered at the location of the object. Its orientation is fixed so that it points at the centroid of the object. A point cloud is obtained at each viewpoint, key points are selected, and local features are extracted (top right). The features are used to construct a VP-Tree (bottom right).}
	\label{fig:GastankViewsphere}
\end{figure}

A simulated depth sensor is used to extract templates from a model by observing it from a discrete set $\{v_1(\rho),\ldots, v_G(\rho)\} \subset V(\rho)$ of viewpoints (Fig. \ref{fig:GastankViewsphere}), which need not be the same as the set of planning viewpoints $\mathcal{X}(\rho)$. The obtained point clouds are collected in a training set $\mathcal{T} := \{\mathcal{P}_{g,l} \mid g = 1,\ldots, G, l = 1,\ldots,|\mathcal{D}|\}$. Features, which describe the local surface curvature are extracted for each template as described below and are used to train the VP-Tree. Given a query point cloud at test time, we extract a set of features and use the VP-Tree to find the template from $\mathcal{T}$, whose features match those of the query the closest.

\subsection{Feature extraction}
\label{subsec:feat_extract}
It is necessary to identify a set of keypoints $\mathcal{K}_\mathcal{P}$ for each template $\mathcal{P} \in \mathcal{T}$, at which to compute local surface features. Since most 3D features are some variation of surface normal estimation, they are very sensitive to noise. As a result, using a unique keypoint estimator is prone to errors. Instead, the keypoints are obtained by sampling the point cloud uniformly (Fig. \ref{fig:GastankViewsphere}), which accounts for global appearance and reduces noise sensitivity. Neighboring points within a fixed radius of every keypoint are used to compute Fast Point Feature Histograms \cite{RusuDoctoralDissertation}. The same number of local features are computed at every keypoint since the radius of the support region is fixed. The features are filtered using a pass-through filter to eliminate badly conditioned ones and are assembled in the set $\{f\}_{kp}$ associated with $kp \in \mathcal{K}_\mathcal{P}$.

\subsection{Training the VP-Tree}
The features $\bigcup_{\mathcal{P} \in \mathcal{T}}\bigcup_{kp \in \mathcal{K}_\mathcal{P}} \{f\}_{kp}$ obtained from the training set are quantized hierarchically into visual words, which are defined by $k$-means clustering (see \cite{NisterS06_vocabtree} for more details). Instead of performing unsupervised clustering, the initial cluster centers are associated with one feature from each of the models in $\mathcal{D}$. The training set $\mathcal{T}$ is partitioned into $|\mathcal{D}|$ groups, where each group consists of the features closest to a particular cluster center. The same process is applied to each group of features, recursively defining quantization cells by splitting each cell into $|\mathcal{D}|$ new parts. The tree is determined level by level, up to some maximum number of levels.

Given a query point cloud $\mathcal{Q}$ at test time, we determine its similarity to a template $\mathcal{P}$ by comparing the paths of their features down the vocabulary tree. The relevance of a feature at node $i$ is determined by a weight:
\[
w_i := \ln\biggl(\frac{|\mathcal{T}|}{\eta_i}\biggr),
\]
where \textit{$\eta_i$} is the number of templates from $\mathcal{T}$ with at least one feature path through node $i$. The weights are used to define a query descriptor $q$ and a template descriptor $d_\mathcal{P}$, with $i$-th component:
\[
q_i := n_i w_i \qquad d_i := m_i w_i,
\]
where $n_i$ and $m_i$ are the number of features of the query and the template, respectively, with a path through node $i$. The templates from $\mathcal{T}$ are ranked according to a relevance score:  
\[
s(q,d_\mathcal{P}) := \biggl\| \frac{d_\mathcal{P}}{\| d_\mathcal{P} \|_1} - \frac{q}{\|q\|_1} \biggr\|_1.
\]
The template with the lowest relevance score is the best matching one to $\mathcal{Q}$.

\subsection{Performance of the VP-Tree}
\begin{figure}[htb]
	\includegraphics[width=\linewidth]{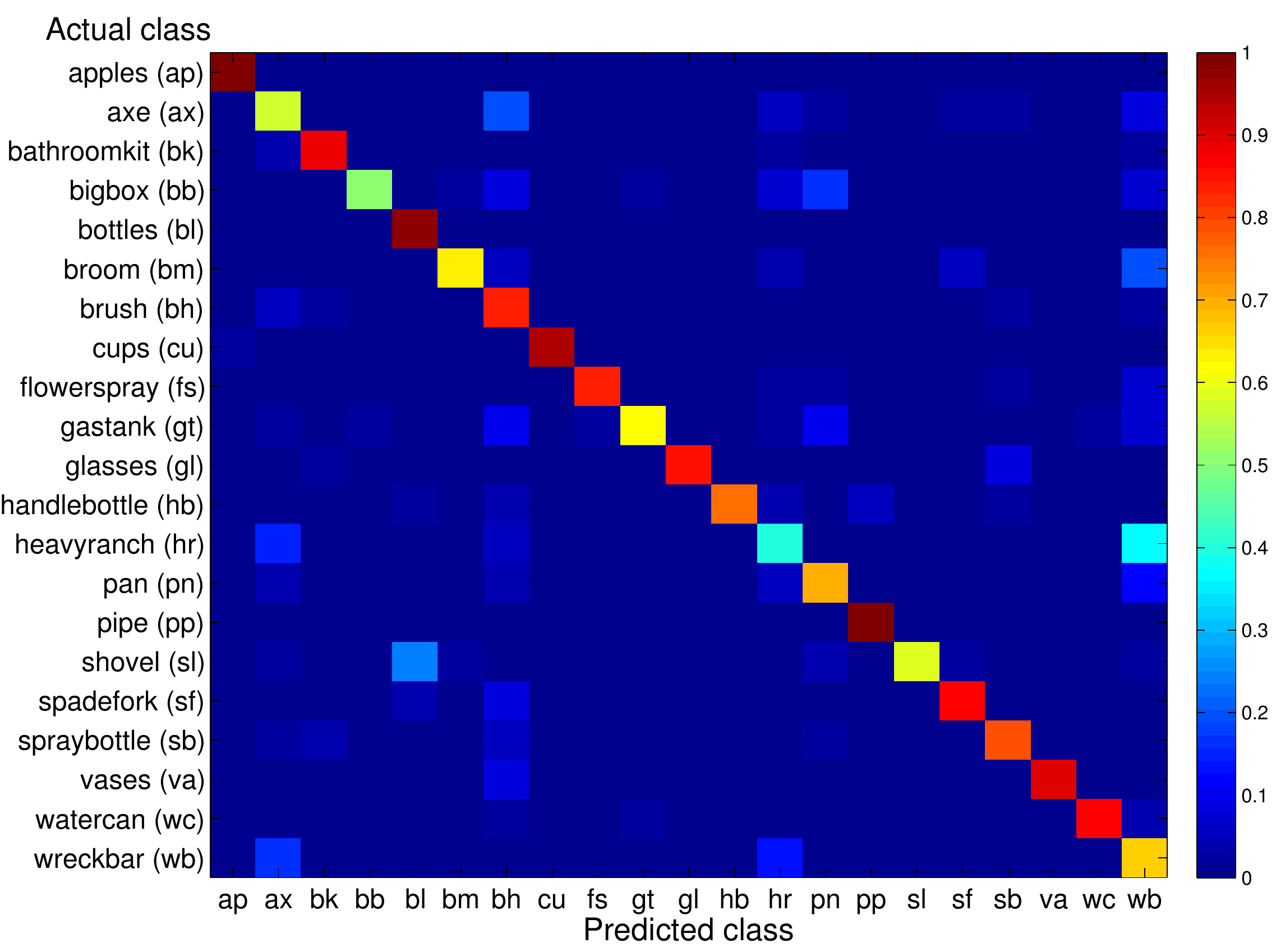}
	\caption{Confusion matrix for all classes in the VP-Tree. A class is formed from all views associated with an object.}
	\label{fig:confusion_result}
\end{figure}

The performance of the static detector was evaluated by using the templates from $\mathcal{T}$ as queries to construct a confusion matrix (Fig. \ref{fig:confusion_result}). If the retrieved template matched the model of the query it was considered correct regardless of the viewpoint.

\begin{figure}[htb]
	\begin{center}
		\includegraphics[width=0.9\linewidth,height=0.45\linewidth,trim=0mm 1mm 0mm 0mm,clip]{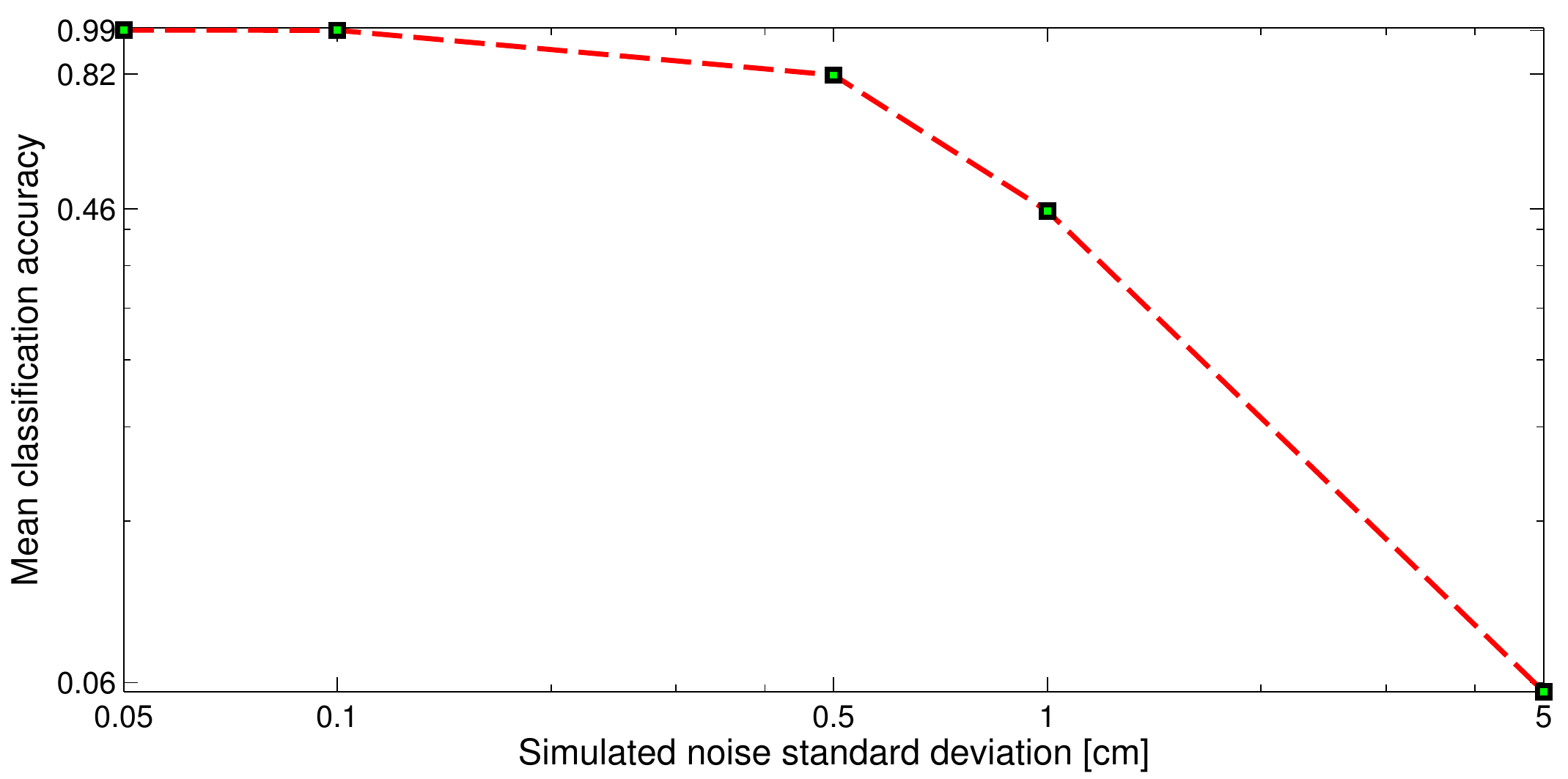}
	\end{center}
	\caption{Effect of signal noise on the classification accuracy of the VP-Tree}
	\label{fig:accuracy_performance}
\end{figure}

To analyze the noise sensitivity of the VP-Tree, we gradually increased the noise added to the test set. Gaussian noise with standard deviation varying from $0.05$ to $5$ cm on a log scale was added along the direction of the ray cast from the observer's viewpoint. The resulting class retrieval accuracy is shown in Fig. \ref{fig:accuracy_performance}. As expected the performance starts to degrade as the amount of noise is increased. However, the detector behaves well at the typical RGB-D camera noise levels.

\section{Observation Model}
\label{sec:obs_model}

This section describes how to obtain statistics about the operation of the sensor from different viewpoints and for different object classes. These statistics are required to maintain a probability distribution over the object hypotheses in a Bayesian framework. Instead of using the segmented point cloud as the observation of the sensor, we take the output of the VP-Tree. This reduces the observation space from all possible point clouds to the space of VP-Tree outputs and includes the operation of the vision algorithm in the sensor statistics. Given a query point cloud suppose that the VP-Tree returns template $\mathcal{P}_{g,l}$ as the top match. Assume that the templates in $\mathcal{T}$ are indexed so that those obtained from models in $\mathcal{I}$ have a lower $l$ index than the rest. We take the linear index of $\mathcal{P}_{g,l}$ as the observation if the match is an object of interest. Otherwise, we record only the model index $l$, ignoring the viewpoint $g$:
\[
Z = \begin{cases}
(l-1)G+g, & \text{if } l \leq |\mathcal{I}|\\
G|\mathcal{I}|+(l-|\mathcal{I}|), & \text{if } l > |\mathcal{I}|.
\end{cases}
\]
This makes the observation space $\mathcal{Z}$ one dimensional.

In order to compute the likelihood of an observation off-line, we introduce an occlusion state $\psi$ for a point cloud. Suppose that the $z$-axis in the sensor frame measures depth and the $xy$-plane is the image plane. Given parameters $\epsilon$ and $\mathcal{E}$, we say that a point cloud is occluded from left if it has less than $\mathcal{E}$ points in the image plane to the left of the line $x = -\epsilon$. If it has less than $\mathcal{E}$ points in the image plane above the line $y = \epsilon$, it is occluded from top. Similarly, we define occluded from bottom, occluded from right, and combinations of them (left-right, left-top, etc.). Let $\Psi$ denote the set of occlusion states, including the non-occluded ($\psi_\emptyset$) and the fully-occluded cases. Then, the data likelihood of an observation $z$ for a given sensor pose $x \in \mathcal{X}(\rho)$, hypothesis $H(c,r)$, and occlusion $\psi\in\Psi$ is:
\[
h_z(x, c, r, \psi) := \mathbb{P}(Z = z \mid x, H(c,r), \psi)
\]
The function $h$ is called the \textit{observation model} of the static detector. It can be obtained off-line because for a given occlusion state it only depends on the characteristics of the sensor and the vision algorithm. Since all variables are discrete, $h$ can be represented with a histogram, which we compute from the training set $\mathcal{T}$. 

\begin{figure*}[t]
	\begin{center}
		\includegraphics[width=0.75\linewidth]{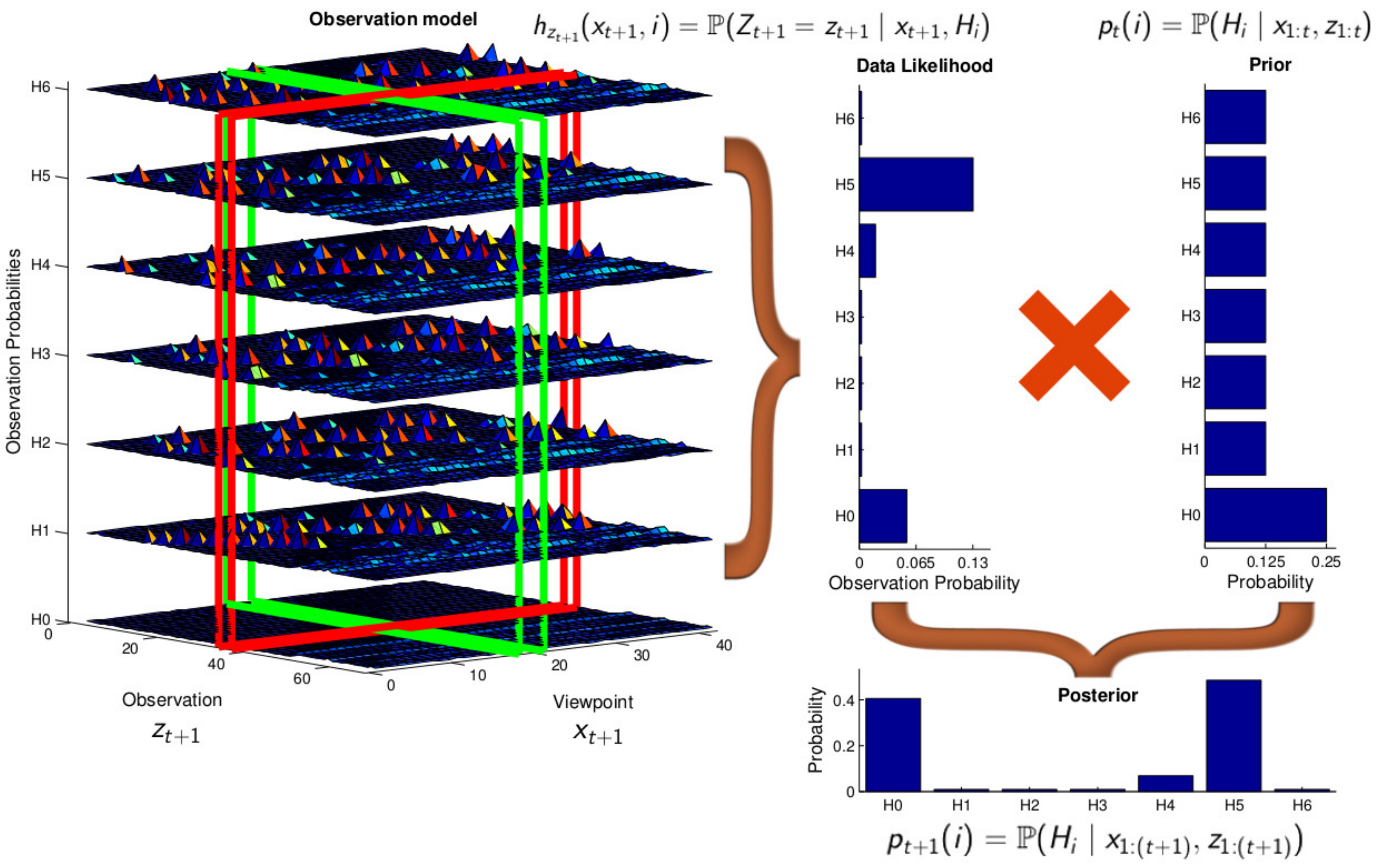}
	\end{center}
	\caption{Observation model obtained with seven hypotheses for the Handlebottle model and the planning viewpoints used in the simulation experiments (Subsection \ref{subsec:static_v_active}). Given a new VP-Tree observation ($z_{t+1}$) from the next viewpoint ($x_{t+1}$), the observation model is used to determine the data likelihood of the observation and to update the prior by applying Bayes rule.}
\label{fig:BayesRule}	
\end{figure*}

Note, however, that the observation model depends on the choice of planning viewpoints and hypotheses, which means that it needs to be recomputed if they change. We would like to compute it once for a given training set and then be able to handle scenarios with different sets of hypotheses and different planning viewpoints. The viewsphere $V(\rho)$ is discretized very finely into a new set of viewpoints $V^o(\rho)$ with coordinates in the frame of the objects in $\mathcal{D}$. A nominal observation model:
\[
  h_z^o(v,c,\psi) := \mathbb{P}(Z = z \mid v, c, \psi), \quad v \in V^o(\rho),c \in \mathcal{D}, \psi \in \Psi
\]
is computed. To obtain $h_z(x,c,r,\psi)$ from the nominal observation model:
\begin{enumerate}
	\item Determine the pose of the sensor $w(x,r)$ in the object frame of $c$.
	\item Find the closest viewpoint $v \in V^o(\rho)$ to $w(x,r)$ (the fine discretization avoids a large error).
	\item Determine the new occlusion region, i.e. rotate the lines associated with $\psi$, in the object frame of $c$. Obtain a point cloud from $v$, remove the points within the occlusion region, and determine the occlusion state $\psi^o$ in the object frame. 
  \item Copy the values from the nominal observation model:
	\[
	  h_z(x,c,r,\psi) = h_z^o(v,c,\psi^o)
	\] 
\end{enumerate}
As a result, it is necessary to compute only the nominal observation model $h_z^o(v,c,\psi^o)$. The histogram representing $h^o$ was obtained in simulation. A viewsphere with radius $\rho = 1m$ was discretized uniformly into $128$ viewpoints (the set $V^o(\rho)$). A simulated depth sensor was used to obtain $20$ independent scores from the VP-Tree for every viewpoint $v \in V^o(\rho)$, every model $c \in \mathcal{D}$, and every occlusion state $\psi \in \Psi$. Fig. \ref{fig:BayesRule} shows an example of the final observation model obtained from the nominal one with the planning viewpoints and hypotheses used in some of our experiments. 

\section{Active Hypothesis Testing}
\label{sec:hyp_test}
In this section we provide a dynamic programming (DP) formulation for the single object optimization problem in (\ref{eq:sequential_optimization}). To simplify notation let $\bar{\mathcal{I}} := \mathcal{I} \cup \{c_\emptyset\}$ and $M := \sum_{c \in \bar{\mathcal{I}}} |\mathcal{R}(c)|$. The state at time $t$ consists of the sensor pose $x_t \in \mathcal{X}(\rho)$ and the \textit{information state} $p_t \in [0,1]^M$, summarized by the sufficient statistic consisting of the probabilities for each hypothesis:
\[
p_t(c,r) := \mathbb{P}(H(c,r) \mid x_{1:t}, z_{1:t}, \psi_{1:t}) \in [0,1],
\]
where $c \in \bar{\mathcal{I}}$, $r \in \mathcal{R}(c)$, $z_{1:t}$ are the VP-Tree observations, and $\psi_{1:t}$ are the occlusion states of the observed point clouds. Suppose that the sensor decides to continue observing by moving to a new viewpoint $x_{t+1}\in\mathcal{X}(\rho)$. The new point cloud is used to determine the VP-Tree score $z_{t+1}$ and the occlusion state $\psi_{t+1}$. The probabilities of the hypotheses are updated according to Bayes' rule:
\begin{align*}
p_{t+1} &= T(p_t,x_{t+1}, z_{t+1}, \psi_{t+1}), \text{ with $(c,r)$ component:}\\
p_{t+1}(c&,r) = \mathbb{P}(H(c,r) \mid x_{1:(t+1)}, z_{1:(t+1)}, \psi_{1:(t+1)})\\
&=\frac{\mathbb{P}(Z_{t+1} = z_{t+1} \mid x_{t+1}, H(c,r), \psi_{t+1})p_t(c,r)}{\mathbb{P}(Z_{t+1} = z_{t+1}\mid x_{t+1}, \psi_{t+1})}\\
&=\frac{h_{z_{t+1}}(x_{t+1},c,r,\psi_{t+1})p_t(c,r)}{\sum_{c' \in \bar{\mathcal{I}}} \sum_{r' \in \mathcal{R}(c)} h_{z_{t+1}}(x_{t+1},c',r',\psi_{t+1}) p_t(c',r') },
\end{align*}
using the assumption of independence of successive observations. See Fig. \ref{fig:BayesRule} for an example of using the observation model to update the probabilities of the hypotheses.

The future sequence of viewpoints is planned with the assumption that there are no occlusions, i.e. $\psi_s  = \psi_\emptyset$ for $s > t+1$. Supposing that $\tau$ is fixed for a moment, the terminal cost of the dynamic program can be derived after the latest observation $z_\tau$ has been incorporated in the posterior:
\begin{align*}
J_\tau(x_\tau,p_\tau) &= \min_{\hat{c} \in \bar{\mathcal{I}}, \hat{r} \in \mathcal{R}(\hat{c})} \mathbb{E}_{c,r} \{\lambda J_D(\hat{c},\hat{r},c,r)\} \\
&= \min_{\hat{c} \in \bar{\mathcal{I}}, \hat{r} \in \mathcal{R}(\hat{c})} \sum_{c \in \bar{\mathcal{I}}} \sum_{r \in \mathcal{R}(c)} \lambda J_D(\hat{c},\hat{r},c,r) p_\tau(c,r).
\end{align*}
The intermediate stage costs for $t = 0,\ldots,(\tau-1)$ are:
\begin{align*}
J_t(x_t,p_t) = &\min_{v \in \mathcal{X}(\rho)} \biggl\{ g(x_t, v) + \\
&\mathbb{E}_{Z_{t+1}} J_{t+1} (v, T(p_t, v, Z_{t+1}, \psi_\emptyset)) \biggr\}.
\end{align*}
Letting $\tau$ be random again and $t$ go to infinity we get the following infinite-horizon dynamic programming equation:
\begin{align}
\label{eq:mary_hypothesis}
J(x,p) = &\min\biggl\{ \min_{\hat{c} \in \bar{\mathcal{I}}, \hat{r} \in \mathcal{R}(\hat{c})} \sum_{c \in \bar{\mathcal{I}}} \sum_{r \in \mathcal{R}(c)} \lambda J_D(\hat{c},\hat{r},c,r) p_\tau(c,r), \notag\\
&\min_{v \in \mathcal{X}(\rho)} g(x,v) + \mathbb{E}_{Z} \{J(v,T(p,v,Z, \psi_\emptyset))\}\biggr\},
\end{align}
which is well-posed by Propositions 9.8 and 9.10 in \cite{Bertsekas07_SoC}. Equation (\ref{eq:mary_hypothesis}) gives an intuition about the relationship between the cost functions $g(\cdot,\cdot)$, $J_D$, and the stopping time $\tau$. If at time $t$, the expected cost of making a mistake given by $\min_{\hat{c} \in \bar{\mathcal{I}}, \hat{r} \in \mathcal{R}(\hat{c})} \sum_{c \in \bar{\mathcal{I}}} \sum_{r \in \mathcal{R}(c)} \lambda J_D(\hat{c},\hat{r},c,r) p_t(c,r)$ is smaller than the cost of taking one more measurement, the sensor stops and chooses the minimizing hypothesis; otherwise it continues measuring.

We resort to numerical approximation techniques, which work well when the state space of the problem is sufficiently small. Define the set $A := \{(c,r) \mid c \in \mathcal{I}, \; r \in \mathcal{R}(c)\} \cup \{(c_\emptyset,r_\emptyset)\}$ of all hypothesized class-orientation pairs. Then, for $s_1, s_2 \in \mathcal{X}(\rho) \cup A$ redefine the cost of movement and the state transition function:
\begin{align*}
g'(s_1,p,s_2) &= \begin{cases}
				g(s_1, s_2), \qquad s_1,s_2 \in \mathcal{X}(\rho) &\\
				\sum_{c \in \bar{\mathcal{I}}} \sum_{r \in \mathcal{R}(c)} \lambda J_D(c',r',c,r) p(c,r), &\\
				\qquad \qquad \qquad s_1 \in \mathcal{X}(\rho), s_2 = (c',r') \in A&\\
				0, \qquad \qquad \quad s_1=s_2 \in A&\\
				\infty, \qquad \qquad \; \text{otherwise}&
			\end{cases}\\
T'(p,s,&z, \psi_\emptyset) = \begin{cases}
				T(p,s,z, \psi_\emptyset), & s \in \mathcal{X}(\rho)\\
				p, & s \in A
			\end{cases}
\end{align*}
We can rewrite (\ref{eq:mary_hypothesis}) into the usual Bellman optimality equation for a POMDP:
\[
J(s,p) = \min_{s' \in \mathcal{X}(\rho)\cup A} \biggl\{g'(s,p,s') + \mathbb{E}_Z\{J(s', T'(p,s',Z,\psi_\emptyset) \} \biggr\}
\]
The state space of the resulting DP is the discrete space of sensor poses $\mathcal{X}(\rho)$ and the continuous space $\mathcal{B} := [0,1]^M$ of distributions over the $M$ hypotheses. Since the viewpoints are chosen locally around the object, the space $\mathcal{X}(\rho)$ is very small in practice (only $42$ viewpoints were used in our experiments). The main computational challenge comes from the dimension $M$ of the continuous space. The size of $\mathcal{B}$ grows exponentially with the number of hypotheses. To alleviate this difficulty, we apply a point-based POMDP algorithm \cite{Kurniawati08_sarsop, Ong08_sarsop}, which uses samples to compute successive approximations to the optimally reachable part of $\mathcal{B}$. An approximate stationary policy $\hat{\mu}: \biggl(\mathcal{X}(\rho)\cup A\biggr) \times \mathcal{B} \rightarrow \mathcal{X}(\rho)\cup A$ is obtained.

In practice, there is some control over the size of $M$. In most applications, the number of objects of interest is small and we show in Subsection \ref{subsec:orient_accuracy} that a very sparse discretization of the orientation space is sufficient to obtain accurate orientation estimates.

\section{Implementation Details}
\label{sec:impl_det}
The previous sections developed a procedure for making a decision about the class and pose of a single object. In this section we present the details of using this procedure to process all objects in the scene as quickly as possible.
 
\subsection{Segmentation and data association}
\label{subsec:seg_data_ass}
Our experiments were performed in a tabletop setting, which simplifies the problems of segmentation and data association. The point clouds received from the scene are clustered according to Euclidean distance by using a Kd-tree. An occupancy grid representing the 2D table surface is maintained in order to associate the clustered surfaces with new or previously seen objects. The centroid of a newly obtained surface is projected to the table and compared with the occupied cells. If the new centroid is close enough to an existing object, the surface is associated with that object and the cell is indexed by the existing object ID. Otherwise, a new object with a unique ID is instantiated.

\subsection{Coupling between objects}
The optimization in Problem \ref{prob:active_obj_detect} is with respect to a single object but while executing it, the sensor obtains surfaces from other objects within its field of view. We have the sensor turn towards the centroid and update the hypotheses' probabilities of every visible object. The turning is required because the observation model was trained only for a sensor facing the centroid of an object. Removing this assumption requires more training data and complicates the observation model computation. The energy used for these turns is not included in the optimization in (\ref{eq:sequential_optimization}). 

The scores obtained from the VP-Tree are not affected significantly by scaling. This allows us to vary the radius $\rho$ of the viewsphere in order to ease the sensor movement and to update hypotheses for other objects within the field of view. The radius is set to $1$ meter by default but if the next viewpoint is not reachable, its can be adapted to accommodate for obstacles and the sensor dynamics. Algorithm \ref{alg:test_pipeline} summarizes the complete view planning framework.

\begin{algorithm}[htb]
\caption{View Planning for Active Object Detection}
\label{alg:test_pipeline}
\begin{algorithmic}[1]
\footnotesize
\State \textbf{Input}: Initial sensor pose $x_1=(x_1^p,x_1^r)\in \mathbb{R}^3 \ltimes SO(3)$, object models of interest $\mathcal{I}$, vector of priors $p_0 \in [0,1]^M$
\State \textbf{Output}: Decision $\hat{c}^i \in \bar{\mathcal{I}}, \hat{r}^i \in \mathcal{R}(\hat{c}^i)$ for every object $i$ in the scene
\State
\State Priority queue $pq \gets \emptyset$
\State Current object ID $i \gets$ unassigned
\For{$t = 1$ to $\infty$}
	\State Obtain a point cloud $\mathcal{Q}_t$ from $x_t$
	\State Cluster $\mathcal{Q}_t$ and update the table occupancy grid
	\For{every undecided object $j$ seen in $\mathcal{Q}_t$}
		\State Rotate the sensor so that $x_t^r$ faces the centroid of $j$
		\State Get viewsphere radius: $\rho \gets \|x_t^p - centroid(j)\|$
		\State Get closest viewpoint: $v^j \gets \argmin_{v \in \mathcal{X}(\rho)} \|x_t^p - v\|$
		\State Obtain a point cloud $\mathcal{Q}^j$
		\State Get VP-Tree score $z^j$ and occlusion state $\psi^j$ from $\mathcal{Q}^j$
		\State Update probabilities for object $j$: $p_t^j \gets T(p_{t-1}^j,v^j,z^j,\psi^j)$ 
		\If{$j \notin pq$}
			\State Insert $j$ in $pq$ according to probability $j \in \mathcal{I}$: $1 - p_t^j(c_\emptyset,r_\emptyset)$
		\EndIf
	\EndFor
	\If{$i$ is unassigned}
		\If{$pq$ is not empty}
			\State $i \gets pq.pop()$
		\Else
			\Comment{All objects seen so far have been processed.}
			\If{whole scene explored}
				\State \textbf{break}
			\Else
				\State Move sensor to an unexplored area and start over
			\EndIf
		\EndIf
	\EndIf
	\State $x_{t+1} \gets \hat{\mu}(v^i,p_t^i)$
	\If{$x_{t+1} == (c,r) \in A$}
		\State $\hat{c}^i \gets c$, $\hat{r}^i \gets r$, $i \gets$ unassigned, Go to line $19$
	\EndIf
	\State Move sensor to $x_{t+1}$
\EndFor
\end{algorithmic}
\end{algorithm}

\section{Performance Evaluation}
\label{sec:experiments}

The VP-Tree was trained on templates extracted using a simulated depth sensor from $48$ viewpoints, uniformly distributed on a viewsphere of radius $\rho = 1$m (Fig. \ref{fig:GastankViewsphere}). To simplify segmentation and data association our experiments were carried out in a tabletop setting. We used $|\mathcal{X}(\rho)| = 42$ planning viewpoints in the upper hemisphere of the viewsphere to avoid placing the sensor under the table. The following costs were used in all experiments:
\begin{align*}
J_D(\hat{c},\hat{r},c,r) &= \begin{cases}
            0, & \hat{c} = c \text{ and } \hat{r} = r\\
						75, & \text{otherwise}\\
					\end{cases}\\
g(x,x') &= gcd(x,x') + g_0,
\end{align*}
where $gcd(\cdot,\cdot)$ is the great-circle distance between two viewpoints $x, x' \in \mathcal{X}(\rho)$ and $g_0 = 1$ is the measurement cost.

\subsection{Performance evaluation in simulation}
\label{subsec:static_v_active}
A single object of interest (Handlebottle) was used: $\mathcal{I} = \{c_H\}$. Keeping the pitch and roll zero, the space of object yaws was discretized into $6$ bins to formulate hypotheses about the detections:
\begin{align*}
H(\emptyset) := H(c_\emptyset,r_\emptyset) =& \text{ The object is \textit{not} a Handlebottle}\\
H(r) := H(c_H,r) =& \text{ The object is a Handlebottle with yaw }\\
   & \; r \in \{0\degree, 60\degree, 120\degree, 180\degree, 240\degree, 300\degree\}
\end{align*}

Seventy synthetic scenes were generated with $10$ true positives for each of the seven hypotheses. The true positive object was placed in the middle of the table, while the rest of the objects served as occluders. See Fig. \ref{fig:Database} for an example of a simulated scene.

Four approaches for selecting sequences of viewpoints from $\mathcal{X}(\rho)$ were compared. The static approach takes a single measurement from the starting viewpoint and makes a decision based on the output from the VP-Tree. This is the traditional approach in machine perception.

The random approach is a random walk on the viewsphere, which avoids revisiting viewpoints. It ranks the viewpoints, which have not been visited yet, according to the great-circle distance from the current viewpoint. Then, it selects a viewpoint at random among the closest ones. The observation model is used to update the hypotheses' probabilities over time. A heuristic stopping rule is used for this method. The experiment is terminated when the probability of one hypothesis is above 60\%, i.e. $\tau = \inf\{t \geq 0 \mid \exists (c,r) \in A \text{ such that } p_t(c,r) \geq 0.6\}$, and that hypothesis is chosen as the sensor's decision.

The greedy mutual information (GMI) approach is the most widely used approach for sensor management tasks \cite{karasev12_visual_learning, ChaMicKum2012}. Specialized to our setting, the GMI policy takes the following form:
\begin{align*}
\mu_{GMI}&(x,p) = \argmax_{x' \in \mathcal{NV}} \frac{\mathbf{I}(H(c,r);Z)}{g(x,x')}\\
&= \argmin_{x' \in \mathcal{NV}} \frac{\mathbf{H}(H(c,r) \mid Z)}{g(x,x')}\\ 
&= \argmin_{x' \in \mathcal{NV}} \frac{1}{g(x,x')} \sum_{z \in \mathcal{Z}} \sum_{c \in \bar{\mathcal{I}}} \sum_{r \in \mathcal{R}(c)} p(c,r) h_z(x',c,r,\psi_\emptyset)\\
&\times \log_2 \biggl(\frac{ \sum_{c' \in \bar{\mathcal{I}}} \sum_{r' \in \mathcal{R}(c')} p(c',r') h_z(x',c',r',\psi_\emptyset)}{p(c,r) h_z(x',c,r,\psi_\emptyset)} \biggr),
\end{align*}
where $\mathcal{NV} := \{x \in \mathcal{X}(\rho) \mid x \text{ has not been visited}\}$, $H(c,r)$ is the true hypothesis, $\mathbf{I}(\cdot;\cdot)$ is mutual information, $\mathbf{H}(\cdot \mid \cdot)$ is conditional entropy, and $\mathcal{Z}$ is the space of observations as defined in Section \ref{sec:obs_model}. The same heuristic stopping rule as for the random approach was used.

The last approach is our nonmyopic view planning (NVP) approach. Fifty repetitions with different starting sensor poses were carried out on every scene. For each hypothesis, the measurement cost $\sum_{t=1}^\tau g_0$, the movement cost $\sum_{t=2}^\tau gcd(x_t, x_{t-1})$, and the decision cost $J_D$ were averaged over all repetitions. The accuracy of each approach and the average costs are presented in Table \ref{tab:static_sim_results}.

\begin{table*}[htb!]
	\scriptsize
	 \begin{center}
   \caption{Simulation results for a bottle detection experiment}
   \begin{tabular}{|c|c|c|c|c|c|c|c|c|c|c|c|c|c|}
   	\hline
  		\multicolumn{3}{|c|}{\multirow{2}{*}{}} & \multicolumn{7}{c|}{\textbf{True Hypothesis}} & \multirow{2}{*}{\shortstack[c]{Avg Number of\\ Measurements}} & \multirow{2}{*}{\shortstack[c]{Avg Movement\\ Cost}} & \multirow{2}{*}{\shortstack[c]{Avg Decision\\ Cost}} & \multirow{2}{*}{\shortstack[c]{Avg Total\\ Cost}}\\ \cline{4-10}
  		\multicolumn{3}{|c|}{} & H(0\degree) & H(60\degree) & H(120\degree) & H(180\degree) & H(240\degree) & H(300\degree) & H($\emptyset$) &  &  &  &  \\ \cline{1-14}
  		\multirow{36}{*}{\begin{sideways}{\textbf{Predicted Hypothesis (\%)}}\end{sideways}} & 		\multirow{8}{*}{\begin{sideways}{Static}\end{sideways}} & H(0\degree) & \tcn{60.35} & \tcn{3.86} & \tcn{1.00} & \tcn{2.19} & \tcn{1.48} & \tcn{2.19} & \tcn{28.92} & 1.00 & 0.00 & 29.74 & 30.74\\ \cline{3-14}
~ & ~ & H(60\degree) & \tcn{5.53} & \tcn{53.90} & \tcn{2.19} & \tcn{1.00} & \tcn{1.48} & \tcn{1.95} & \tcn{33.94} & 1.00 & 0.00 & 34.57 & 35.57\\ \cline{3-14}
~ & ~ & H(120\degree) & \tcn{4.86} & \tcn{4.62} & \tcn{51.49} & \tcn{3.90} & \tcn{2.21} & \tcn{1.24} & \tcn{31.68} & 1.00 & 0.00 & 36.38 & 37.38\\ \cline{3-14}
~ & ~ & H(180\degree) & \tcn{4.34} & \tcn{4.34} & \tcn{6.01} & \tcn{49.13} & \tcn{1.95} & \tcn{1.24} & \tcn{32.98} & 1.00 & 0.00 & 38.15 & 39.15\\ \cline{3-14}
~ & ~ & H(240\degree) & \tcn{3.88} & \tcn{1.96} & \tcn{1.24} & \tcn{2.20} & \tcn{56.11} & \tcn{1.24} & \tcn{33.37} & 1.00 & 0.00 & 32.92 & 33.92\\ \cline{3-14}
~ & ~ & H(300\degree) & \tcn{5.07} & \tcn{1.24} & \tcn{2.44} & \tcn{2.44} & \tcn{1.72} & \tcn{54.29} & \tcn{32.82} & 1.00 & 0.00 & 34.28 & 35.28\\ \cline{3-14}
~ & ~ & H($\emptyset$) & \tcn{0.56} & \tcn{1.09} & \tcn{3.11} & \tcn{1.93} & \tcn{0.32} & \tcn{3.13} & \tcn{89.87} & 1.00 & 0.00 & 7.60 & 8.60\\ \cline{3-14}
		~ & ~ &\multicolumn{11}{|r|}{Overall Average Total Cost: } & \textbf{\textcolor{red}{31.52}}\\ \cline{2-14}
  		~ & \multirow{8}{*}{\begin{sideways}{Random}\end{sideways}} & H(0\degree) & \tcn{73.78} & \tcn{3.17} & \tcn{1.24} & \tcn{2.21} & \tcn{1.48} & \tcn{1.24} & \tcn{16.87} & 2.00 & 1.26 & 19.66 & 22.93\\ \cline{3-14}
~ & ~ & H(60\degree) & \tcn{1.96} & \tcn{70.34} & \tcn{2.20} & \tcn{1.72} & \tcn{1.00} & \tcn{1.48} & \tcn{21.31} & 2.36 & 1.71 & 22.25 & 26.31\\ \cline{3-14}
~ & ~ & H(120\degree) & \tcn{1.00} & \tcn{1.49} & \tcn{70.75} & \tcn{3.43} & \tcn{1.00} & \tcn{1.24} & \tcn{21.09} & 2.30 & 1.64 & 21.94 & 25.87\\ \cline{3-14}
~ & ~ & H(180\degree) & \tcn{1.48} & \tcn{1.73} & \tcn{3.66} & \tcn{66.97} & \tcn{1.97} & \tcn{1.48} & \tcn{22.71} & 2.71 & 2.16 & 24.78 & 29.64\\ \cline{3-14}
~ & ~ & H(240\degree) & \tcn{1.48} & \tcn{1.24} & \tcn{1.48} & \tcn{2.45} & \tcn{68.76} & \tcn{1.72} & \tcn{22.87} & 2.41 & 1.77 & 23.43 & 27.62\\ \cline{3-14}
~ & ~ & H(300\degree) & \tcn{1.72} & \tcn{1.97} & \tcn{1.00} & \tcn{1.24} & \tcn{1.97} & \tcn{71.85} & \tcn{20.25} & 2.60 & 2.02 & 21.11 & 25.74\\ \cline{3-14}
~ & ~ & H($\emptyset$) & \tcn{0.07} & \tcn{2.11} & \tcn{2.00} & \tcn{1.53} & \tcn{1.59} & \tcn{0.37} & \tcn{92.33} & 4.95 & 4.93 & 5.76 & 15.64\\ \cline{3-14}
		~ & ~ &\multicolumn{11}{|r|}{Overall Average Total Cost:} & \textbf{\textcolor{red}{24.82}} \\ \cline{2-14}
  		~ & \multirow{8}{*}{\begin{sideways}{Greedy MI}\end{sideways}} & H(0\degree) & \tcn{82.63} & \tcn{2.93} & \tcn{0.76} & \tcn{1.61} & \tcn{0.83} & \tcn{0.40} & \tcn{10.85} & 1.96 & 1.20 & 13.03 & 16.19\\ \cline{3-14}
~ & ~ & H(60\degree) & \tcn{0.80} & \tcn{80.14} & \tcn{1.05} & \tcn{1.07} & \tcn{0.14} & \tcn{1.16} & \tcn{15.64} & 2.26 & 1.58 & 14.89 & 18.73\\ \cline{3-14}
~ & ~ & H(120\degree) & \tcn{1.09} & \tcn{1.05} & \tcn{76.93} & \tcn{2.64} & \tcn{0.83} & \tcn{0.82} & \tcn{16.66} & 2.30 & 1.64 & 17.31 & 21.25\\ \cline{3-14}
~ & ~ & H(180\degree) & \tcn{1.47} & \tcn{1.25} & \tcn{3.62} & \tcn{75.60} & \tcn{0.71} & \tcn{0.50} & \tcn{16.84} & 2.79 & 2.25 & 18.30 & 23.34\\ \cline{3-14}
~ & ~ & H(240\degree) & \tcn{0.49} & \tcn{1.15} & \tcn{0.82} & \tcn{2.58} & \tcn{75.29} & \tcn{1.71} & \tcn{17.96} & 2.37 & 1.72 & 18.53 & 22.62\\ \cline{3-14}
~ & ~ & H(300\degree) & \tcn{1.79} & \tcn{0.50} & \tcn{0.12} & \tcn{0.86} & \tcn{1.21} & \tcn{81.78} & \tcn{13.74} & 2.59 & 2.00 & 13.66 & 18.25\\ \cline{3-14}
~ & ~ & H($\emptyset$) & \tcn{0.72} & \tcn{1.35} & \tcn{2.23} & \tcn{0.39} & \tcn{0.25} & \tcn{0.41} & \tcn{94.65} & 5.29 & 5.37 & 4.01 & 14.67\\ \cline{3-14}
		~ & ~ & \multicolumn{11}{|r|}{Overall Average Total Cost:} & \textbf{\textcolor{red}{19.29}} \\ \cline{2-14}
  		~ & \multirow{8}{*}{\begin{sideways}{NVP}\end{sideways}} & H(0\degree) & \tcn{87.98} & \tcn{0.48} & \tcn{0.24} & \tcn{0.24} & \tcn{0.24} & \tcn{0.48} & \tcn{10.34} & 2.06 & 1.45 & 9.01 & 12.51\\ \cline{3-14}
~ & ~ & H(60\degree) & \tcn{0.00} & \tcn{83.78} & \tcn{0.97} & \tcn{0.24} & \tcn{0.24} & \tcn{0.24} & \tcn{14.53} & 2.28 & 1.73 & 12.17 & 16.17\\ \cline{3-14}
~ & ~ & H(120\degree) & \tcn{0.48} & \tcn{0.00} & \tcn{82.81} & \tcn{1.21} & \tcn{0.00} & \tcn{0.00} & \tcn{15.50} & 2.37 & 1.86 & 12.89 & 17.12\\ \cline{3-14}
~ & ~ & H(180\degree) & \tcn{0.00} & \tcn{0.00} & \tcn{0.97} & \tcn{82.61} & \tcn{1.21} & \tcn{0.24} & \tcn{14.98} & 2.50 & 2.05 & 13.04 & 17.60\\ \cline{3-14}
~ & ~ & H(240\degree) & \tcn{0.49} & \tcn{0.24} & \tcn{0.00} & \tcn{0.49} & \tcn{78.73} & \tcn{0.00} & \tcn{20.05} & 2.57 & 2.18 & 15.95 & 20.71\\ \cline{3-14}
~ & ~ & H(300\degree) & \tcn{0.00} & \tcn{0.24} & \tcn{0.24} & \tcn{0.73} & \tcn{0.48} & \tcn{81.60} & \tcn{16.71} & 2.60 & 2.15 & 13.80 & 18.55\\ \cline{3-14}
~ & ~ & H($\emptyset$) & \tcn{1.49} & \tcn{1.58} & \tcn{1.37} & \tcn{0.37} & \tcn{0.74} & \tcn{1.25} & \tcn{93.20} & 2.08 & 1.50 & 5.10 & 8.68\\ \cline{3-14}
~ & ~ & \multicolumn{11}{|r|}{Overall Average Total Cost:} & \textbf{\textcolor{red}{15.91}} \\\hline
  \end{tabular}
  \end{center}
\label{tab:static_sim_results}
\end{table*}

Compared with the random and the GMI approaches, our method needs less movement and less measurements on average. This can be explained by the fact that the stopping time for NVP is included in the optimization, while for random and GMI it is chosen heuristically. The results also suggest that our approach is able to select more informative viewpoints since it has a lower average decision cost than the other methods. We can conclude that the nonmyopic view planning approach outperforms the widely-used greedy techniques and provides a significant improvement over the traditional static detection.

\subsection{Accuracy of the orientation estimates}
\label{subsec:orient_accuracy}
Since the object orientations in a real scene are not discretized a refinement step is needed if the algorithm detects an object of interest, i.e. decides on $\hat{c} \neq c_\emptyset$. The surfaces observed from an object are accumulated over time. After a decision, these surfaces are aligned using an iterative closest point algorithm with the surface of the database model, corresponding to $H(\hat{c},\hat{r})$. Thus, the final decision includes both a class and a continuous pose estimate.

A set of simulations was carried out in order to evaluate the accuracy of the continuous orientation estimates with respect to the ground truth. The following distance metric on $SO(3)$ was used to measure the error between two orientations represented by quaternions $q_1$ and $q_2$:
\[
d(q_1,q_2) = \cos^{-1}\bigl(2 \langle q_1, q_2 \rangle^2 - 1\bigr),
\]
where $\langle a_1 + b_1\mathbf{i}+c_1\mathbf{j}+d_1\mathbf{k}, a_2 + b_2\mathbf{i}+c_2\mathbf{j}+d_2\mathbf{k} \rangle = a_1a_2 + b_1b_2+c_1c_2 + d_1d_2$ denotes the quaternion inner product.

\begin{figure}[ht!]
	\begin{center}
		\includegraphics[width=\linewidth,trim=11mm 1mm 11mm 15mm,clip]{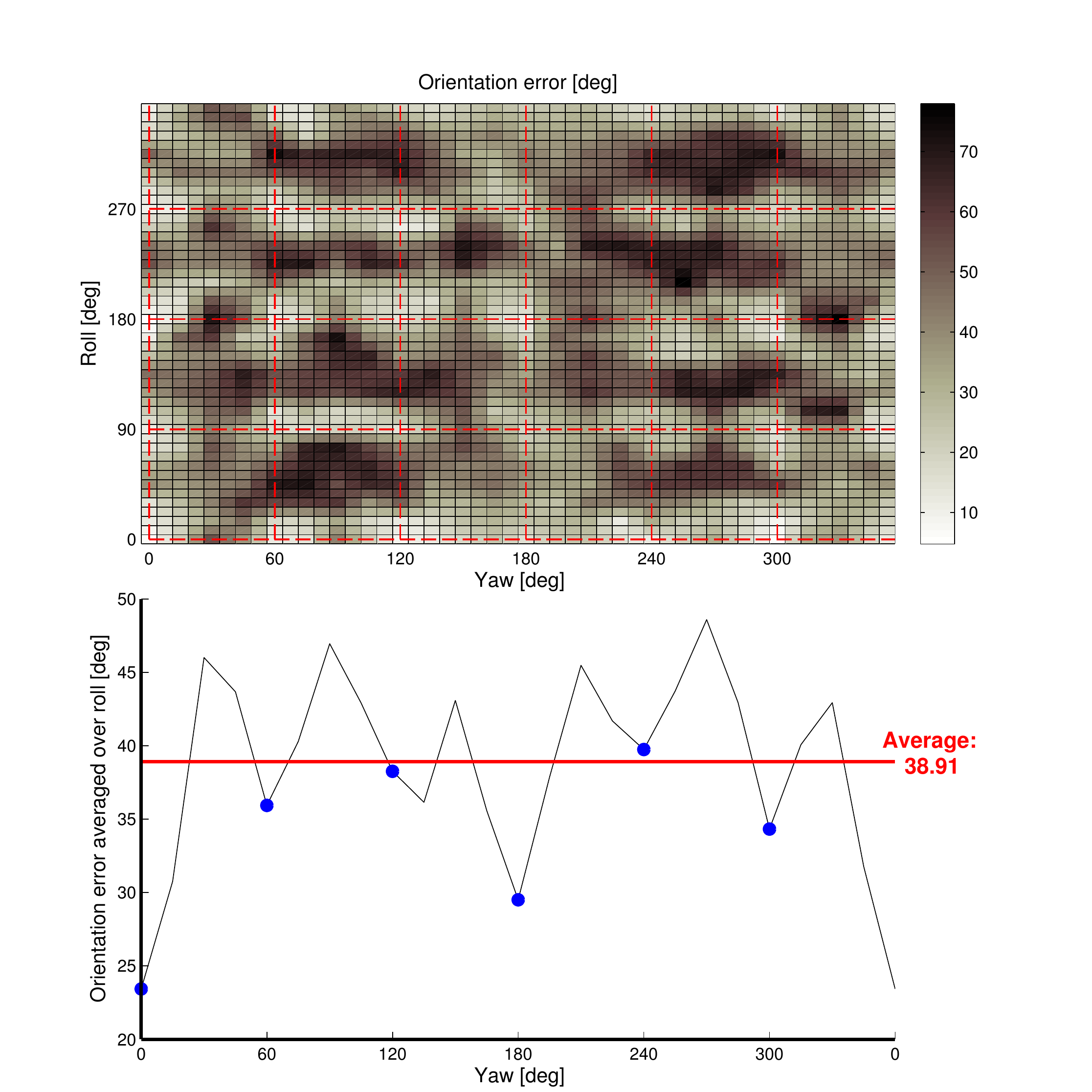}
	\end{center}
	\caption{Twenty five hypotheses (red dotted lines) were used to decide on the orientation of a Watercan. The error in the orientation estimates is shown as the ground truth orientation varies (top). The error averaged over the ground truth roll, the hypotheses over the object yaw (blue dots), and the overall average error (red line) are shown in the bottom plot.}
	\label{fig:orr_error}
\end{figure}

A single object of interest (Watercan) was used: $\mathcal{I} = \{c_W\}$. The ground truth yaw ($\alpha$) and roll ($\gamma$) of the Watercan were varied from $0\degree$ to $360\degree$ at $7.5\degree$ increments. The pitch ($\beta$) was kept zero. Synthetic scenes were generated for each orientation. To formulate hypotheses about the detection, the yaw space was discretized into $6$ bins and the roll space was discretized into $4$ bins:
\begin{alignat*}{2}
H(c_\emptyset,r_\emptyset) &= \text{ The object is \textit{not} a Watercan}&&\\
H(c_W,r) &= \text{ The object is a Watercan with}&&\text{ orientation}\\
r &= (\alpha,\beta,\gamma) \in \{(i_y60\degree,0,i_r90\degree) \mid &&i_y = 0,\ldots,5,\\
&&&i_r = 0,\ldots,3\}
\end{alignat*}
Fifty repetitions with different starting sensor poses were carried out on every test scene. The errors in the orientation estimates were averaged and the results are presented in Fig. \ref{fig:orr_error}. As expected, the orientation estimates get worse for ground truth orientations which are further away from the hypothesized orientations. On the bottom plot, it can be seen that the hypothesized yaws correspond to local minima in the orientation error. This suggests that the number of hypotheses needs to be increased if a better orientation estimate is desired. Still, a rather sparse set of hypothesized orientations was sufficient to obtain an average error of $39\degree$. For these experiments, the average number of measurements was $2.85$ and the average movement cost was $2.61$.

\subsection{Performance evaluation in real-world experiments}
\begin{figure*}[ht!]
	\begin{center}
		\includegraphics[width=\linewidth]{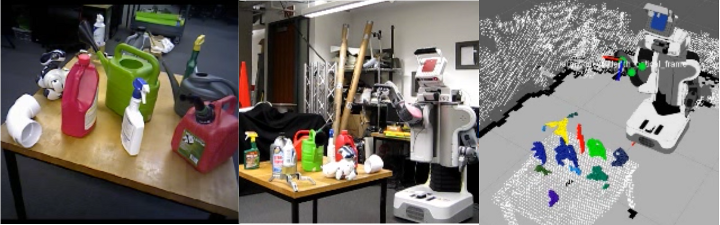}
	\end{center}
	\caption{An example of the experimental setup (left), which contains two instances of the object of interest (Handlebottle). A PR2 robot with an Asus Xtion RGB-D camera attached to the right wrist (middle) employs the nonmyopic view planning approach for active object detection and pose estimation. In the robot's understanding of the scene (right), the object which is currently under evaluation is colored yellow. Once the system makes a decision about an object, it is colored green if it is of interest, i.e. in $\mathcal{I}$, and red otherwise. Hypothesis $H(0\degree)$ (Handlebottle with yaw $0\degree$) was chosen correctly for the green object. See the attached video or \url{http://www.seas.upenn.edu/~atanasov/vid/Atanasov_ActiveObjectDetection_TRO13.mp4} for more details.}
	\label{fig:real_exp}
\end{figure*}

The evaluation of our framework from Subsection \ref{subsec:static_v_active} was recreated in a real environment. An Asus Xtion RGB-D camera attached to the right wrist of a PR2 robot was used as the mobile sensor. As before, the sensor's task was to detect if any Handlebottles ($\mathcal{I} = \{c_H\}$) are present on a cluttered table and estimate their pose. Fig. \ref{fig:real_exp} shows the experimental setup.

Twelve different table setups were used, each containing $2$ instances of the object of interest and $8-10$ other objects. Ten repetitions were carried out for each setup, which in total corresponded to $40$ true positive cases for every hypothesis. The results are summarized in Table \ref{tab:real_results}. The performance obtained in the real experiments is comparable to the simulation results. On average, more movement and more measurements were required to make a decision in practice than in simulation. This can be attributed to the fact that the VP-Tree and the observation model were trained in simulation but were used to process real observations. We expect that better results will be obtained if the training phase is performed using the real sensor with real objects. Still, the results from the experiments are very satisfactory with an average accuracy of $76\%$ for true positives and $98\%$ for true negatives. 

The experiments suggested that several aspects of our framework need improvement. First, the occlusion model should be used in the planning stage to avoid visiting viewpoints with limited visibility. Second, the observation model can be modified, at the expense of a more demanding training stage, to include sensor poses which do not face the object's centroid. This will eliminate the need to turn the sensor towards the centroid of every visible object. As far as computation time is concerned, the main bottleneck was the the feature extraction from the observed surfaces and the point cloud registration needed to localize the sensor in the global frame (our method assumes that the sensor has accurate self-localization). To demonstrate that our approach can handle more complicated scenarios, several experiments were performed with two objects of interest (Handlebottle and Watercan): $\mathcal{I} = \{c_H,c_W\}$, and $53$ hypotheses associated with likely poses for the two objects. See the video from Fig. \ref{fig:real_exp} for more details.

\begin{table*}[htb!]
  \scriptsize
  \begin{center}
	\caption{Results for a real-world bottle detection experiment}
	\begin{tabular}{|c|c|c|c|c|c|c|c|c|c|c|c|c|}
   	\hline
  		\multicolumn{2}{|c|}{\multirow{2}{*}{}} & \multicolumn{7}{c|}{\textbf{True Hypothesis}} & \multirow{2}{*}{\shortstack[c]{Avg Number of\\ Measurements}} & \multirow{2}{*}{\shortstack[c]{Avg Movement\\ Cost}} & \multirow{2}{*}{\shortstack[c]{Avg Decision\\ Cost}} & \multirow{2}{*}{\shortstack[c]{Avg Total\\ Cost}}\\ \cline{3-9}
  		\multicolumn{2}{|c|}{} & H(0\degree) & H(60\degree) & H(120\degree) & H(180\degree) & H(240\degree) & H(300\degree) & H($\emptyset$) &  &  &  &  \\ \cline{1-13}
  		\multirow{8}{*}{\begin{sideways}{\textbf{Predicted (\%)}}\end{sideways}} & H(0\degree) & \tcn{87.5} & \tcn{2.5} & \tcn{5.0} & \tcn{0.0} & \tcn{0.0} & \tcn{0.0} & \tcn{5.0} & 2.53 & 2.81 & 9.38 & 14.72\\ \cline{2-13}
~ & H(60\degree) & \tcn{2.5} & \tcn{80.0} & \tcn{0.0} & \tcn{0.0} & \tcn{0.0} & \tcn{0.0} & \tcn{17.5} & 2.66 & 2.52 & 15.00 & 20.18\\ \cline{2-13}
~ & H(120\degree) & \tcn{7.5} & \tcn{0.0} & \tcn{72.5} & \tcn{0.0} & \tcn{0.0} & \tcn{0.0} & \tcn{20.0} & 3.16 & 3.43 & 20.63 & 27.22\\ \cline{2-13}
~ & H(180\degree) & \tcn{0.0} & \tcn{0.0} & \tcn{0.0} & \tcn{70.0} & \tcn{10.0} & \tcn{2.5} & \tcn{17.5} & 2.20 & 1.72 & 22.5 & 26.42\\ \cline{2-13}
~ & H(240\degree) & \tcn{0.0} & \tcn{0.0} & \tcn{0.0} & \tcn{2.5} & \tcn{75.0} & \tcn{2.5} & \tcn{20.0} & 2.39 & 2.51 & 18.75 & 23.65\\ \cline{2-13}
~ & H(300\degree) & \tcn{0.0} & \tcn{0.0} & \tcn{0.0} & \tcn{0.0} & \tcn{5.0} & \tcn{72.5} & \tcn{22.5} & 2.57 & 2.18 & 20.63 & 25.38\\ \cline{2-13}
~ & H($\emptyset$) & \tcn{0.0} & \tcn{0.0} & \tcn{0.97} & \tcn{0.0} & \tcn{0.0} & \tcn{0.97} & \tcn{98.05} & 2.17 & 1.93 & 1.46 & 5.56\\ \cline{2-13}
~ &\multicolumn{11}{|r|}{Overall Average Total Cost: } & \textbf{\textcolor{red}{20.45}}\\\hline	
	\end{tabular}
	\end{center}
	\label{tab:real_results}
\end{table*}

\section{Conclusion}
This work considered the problem of detection and pose estimation of semantically important objects versus background using a depth camera. A novel static detector, the VP-Tree, which combines detection and pose estimation for 3D objects was introduced. To alleviate the difficulties associated with single-view recognition, we formulated hypotheses about the class and orientation of an unknown object and proposed a soft detection strategy, in which the sensor moves to increase its confidence in the correct hypothesis. Non-myopic view planning was used to balance the amount of energy spent for sensor motion with the benefit of decreasing the probability of an incorrect decision.

The validity of our approach was verified both in simulation and in real-world experiments with an Asus Xtion RGB-D camera attached to the wrist of a PR2 robot. Careful analysis was performed in simulation to demonstrate that our approach outperforms the widely-used greedy viewpoint selection methods and provides a significant improvement over the traditional static detection. The simulation experiments were recreated in a real setting and the results show that the performance is comparable to the simulations. 

Our approach has several advantages over existing work. The observation model described in Section \ref{sec:obs_model} is general and applicable to real sensors. The proposed planning framework is independent of the static object detector and can be used with various existing algorithms in machine perception. Finally, instead of using an information theoretic cost function, the probability of an incorrect decision is minimized directly.

The drawback of our approach is that it requires an accurate estimate of the sensor pose and contains no explicit mechanism to handle occlusions during the planning stage. Moreover, the sequence of viewpoints is selected with respect to a single object instead of all objects within the field of view.

In future work, we would like to improve the occlusion model and use it during the planning stage. This will necessitate re-planning as the motion policy will no longer be computable off-line. We would like to obtain a sub-optimal policy, which is simple to re-compute but is still non-myopic. The effect of introducing sensor dynamics in the active M-ary hypothesis testing problem is of great interest as well.




\section*{Acknowledgment}
The authors would like to thank Thomas Koletchka and Ben Cohen for their help with ROS and the PR2.


\ifCLASSOPTIONcaptionsoff
  \newpage
\fi

\bibliographystyle{IEEEtran}
\bibliography{bib/TRO13_Bibliography}
\begin{IEEEbiography}[{\includegraphics[width=1in,height=1.25in,clip,keepaspectratio]{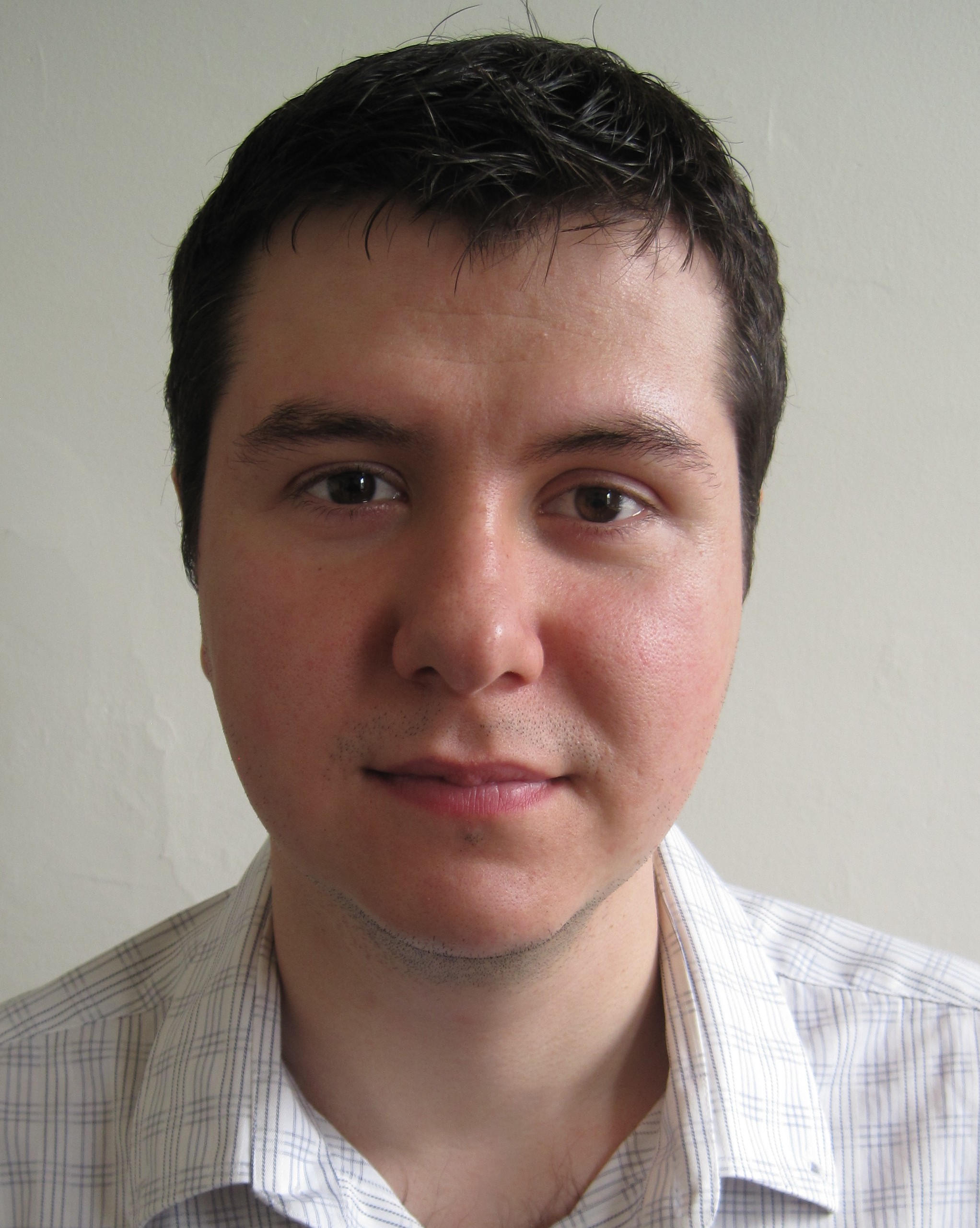}}]%
{Nikolay Atanasov}
(S'07) received the B.S. degree in electrical engineering from Trinity College, Hartford, CT, in 2008 and the M.S. degree in electrical engineering from the University of Pennsylvania, Philadelphia, PA, in 2012, where he is currently working towards the Ph.D. degree in electrical and systems engineering.

His research interests include computer vision with applications to robotics, planning and control of mobile sensors, sensor management, detection and estimation theory.
\end{IEEEbiography}

\begin{IEEEbiography}[{\includegraphics[width=1in,height=1.25in,clip,keepaspectratio]{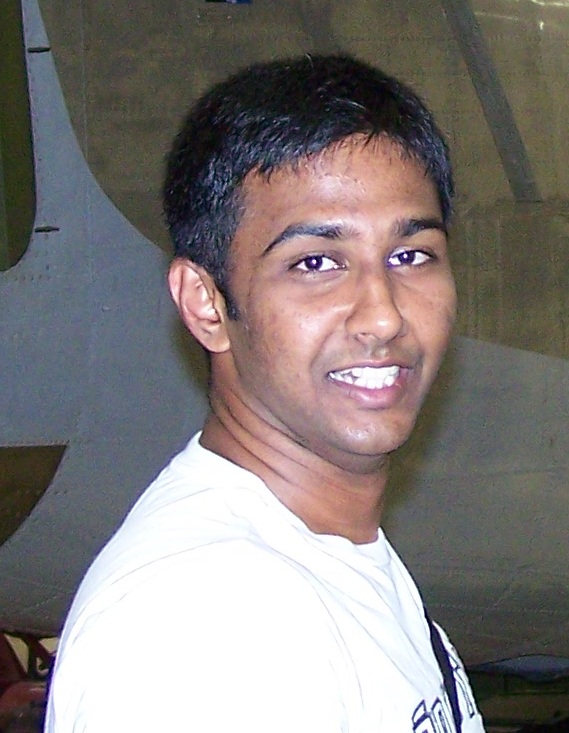}}]%
{Bharath Sankaran}
(S'09) received his B.E degree in mechanical engineering from Anna University, Chennai, India in 2006, an M.E. degree in aerospace engineering from the University of Maryland, College Park, MD, in 2008 and an M.S. degree in robotics from the University of Pennsylvania, Philadelphia, PA, in 2012. He is currently working towards his PhD in computer science at the University of Southern California.

His research interests include applying statistical learning techniques to perception and control problems in robotics, addressing perception-action loops and biped locomotion.
\end{IEEEbiography}

\begin{IEEEbiography}[{\includegraphics[width=1in,height=1.25in,clip,keepaspectratio]{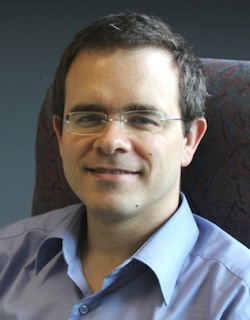}}]%
{Jerome Le Ny}
(S'05-M'09) is an assistant professor in the Department of Electrical Engineering at the \'Ecole Polytechnique de Montr\'eal since May 2012. He received the B.S. degree from the Ecole Polytechnique, France, in 2001, the M.Sc. degree in Electrical Engineering from the University of Michigan, Ann Arbor, in 2003, and the Ph.D. degree in Aeronautics and Astronautics from the Massachusetts Institute of Technology, Cambridge, in 2008. From 2008 to 2012 he was a Postdoctoral Researcher with the GRASP Laboratory at the University of Pennsylvania. His research interests include robust and stochastic control, scheduling and dynamic resource allocation problems, with applications to autonomous and embedded systems, multi-robot systems, and transportation systems.
\end{IEEEbiography}

\begin{IEEEbiography}[{\includegraphics[width=1in,height=1.25in,clip,keepaspectratio]{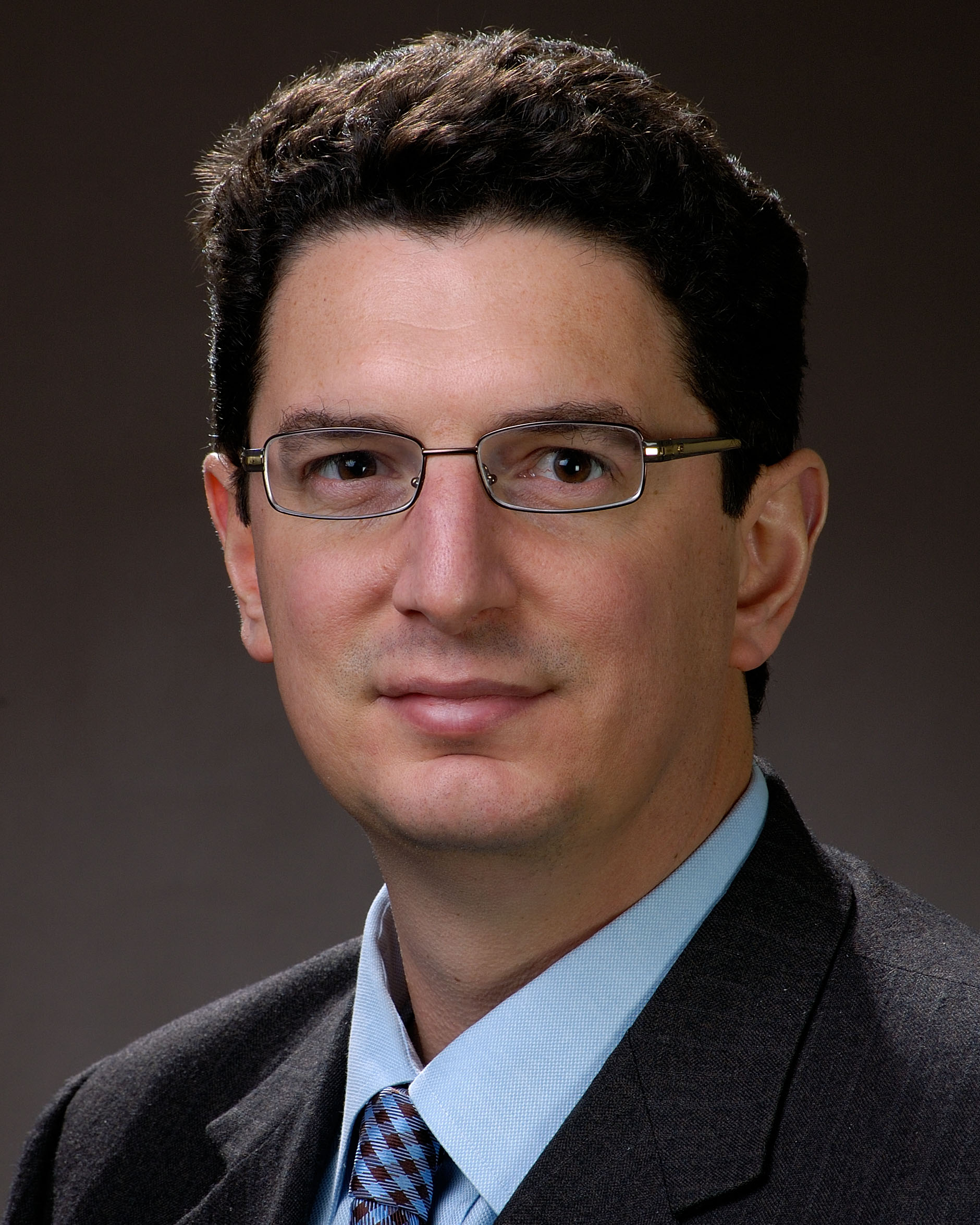}}]%
{George J. Pappas}
(S'90-M'91-SM'04-F'09) received the Ph.D. degree in electrical engineering and computer sciences from the University of California, Berkeley in 1998, for which he received the Eliahu
Jury Award for Excellence in Systems Research.

He is currently the Joseph Moore Professor and Chair of the Department of Electrical and Systems Engineering at the University of Pennsylvania. He also holds a secondary appointment in the Departments of Computer and Information Sciences, and Mechanical Engineering and Applied Mechanics. He is member of the GRASP Lab and the PRECISE Center. He has previously served as the Deputy Dean for Research in the School of Engineering and Applied Science. His research focuses on control theory and in particular, hybrid systems, embedded systems, hierarchical and distributed control systems, with applications to unmanned aerial vehicles, distributed robotics, green buildings, and biomolecular networks. He is a Fellow of IEEE, and has received various awards such as the Antonio Ruberti Young Researcher Prize, the George S. Axelby Award, and the National Science Foundation PECASE.
\end{IEEEbiography}

\begin{IEEEbiography}[{\includegraphics[width=1in,height=1.25in,clip,keepaspectratio]{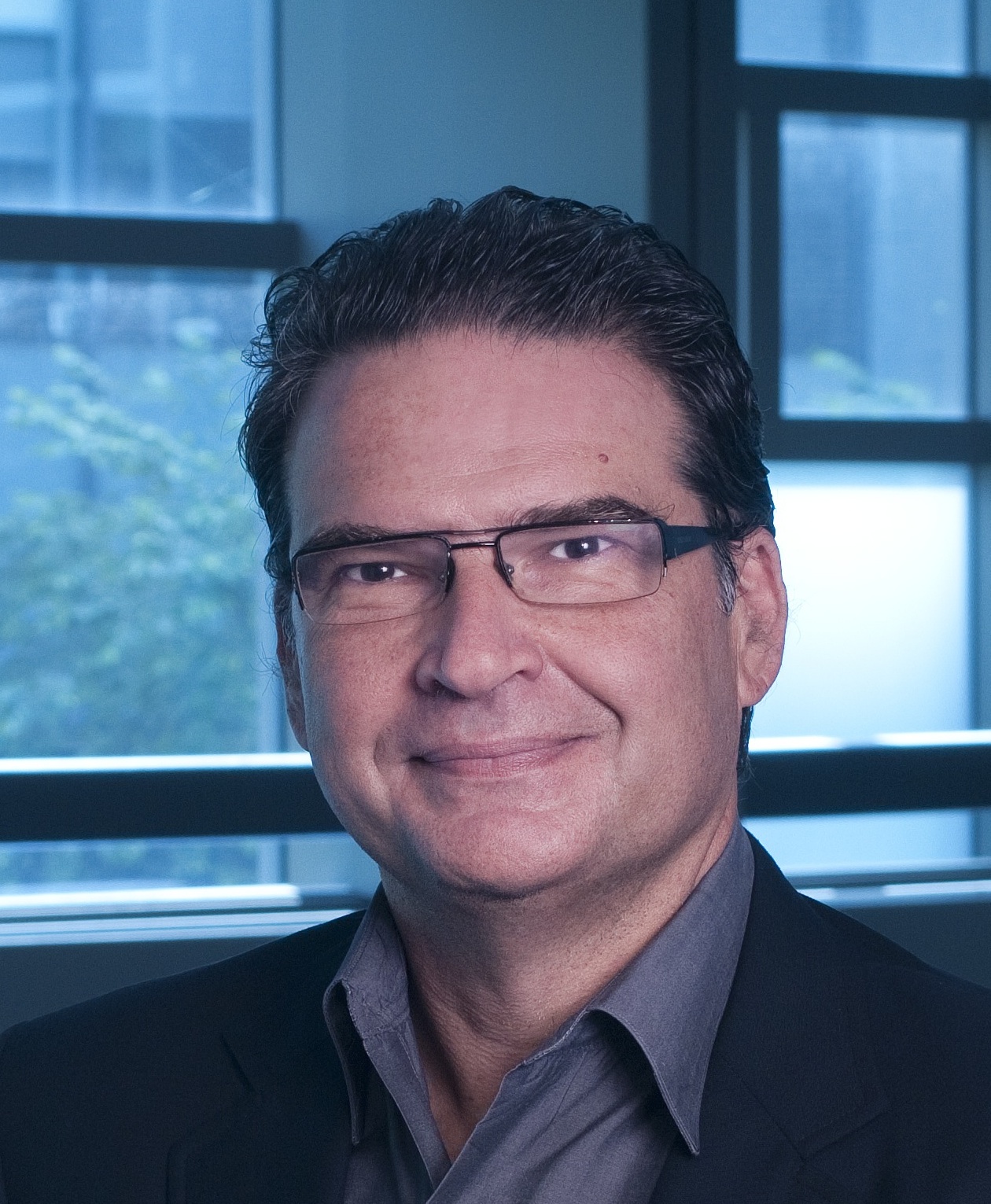}}]%
{Kostas Daniilidis}
(S'90-M'92-SM'04-F'12) is a Professor of Computer and Information Science at the University of Pennsylvania where he has been faculty since 1998. He obtained his undergraduate degree in Electrical Engineering from the National Technical University of Athens, 1986, and his PhD  in Computer Science from the University of Karlsruhe, 1992, under the supervision
of Hans-Hellmut Nagel. His research interests are on visual motion and navigation, image matching, 3D object and place recognition, and camera design. He was Associate Editor of IEEE Transactions on Pattern Analysis and Machine Intelligence from 2003 to 2007. He founded the series of IEEE Workshops on Omnidirectional Vision. In June 2006, he co-chaired with Pollefeys the Third Symposium on 3D Data Processing, Visualization, and Transmission, and he was Program Chair of the 11th European Conference on Computer Vision in 2010. He has been the director of the interdisciplinary GRASP laboratory from 2008 to 2013 and he is serving as the Associate Dean for Graduate Education of Penn Engineering since 2013.
\end{IEEEbiography}

\vfill






\end{document}